\documentclass[10pt,twocolumn,letterpaper]{article}
\usepackage{abstract}
\usepackage{cvpr}
\usepackage{times}
\usepackage{epsfig}
\usepackage{graphicx}
\usepackage{amsmath}
\usepackage{amssymb}
\usepackage{xcolor}
\usepackage{wrapfig}
\usepackage{lipsum}
\usepackage{mathtools}
\usepackage{cuted}
\usepackage{subcaption}
\usepackage{booktabs}
\usepackage{caption}
\usepackage[affil-it]{authblk}
\usepackage{nopageno}

\newcommand{\nothing}[1]{}
\DeclarePairedDelimiter{\ceil}{\lceil}{\rceil}

\usepackage[pagebackref=true,breaklinks=true,letterpaper=true,colorlinks,bookmarks=false]{hyperref}

\newcommand*\samethanks[1][\value{footnote}]{\footnotemark[#1]}

\cvprfinalcopy 

\def\cvprPaperID{4724} 

\ifcvprfinal\fi
\begin{document}


\title{Learning Formation of Physically-Based Face Attributes}

\author{Ruilong Li\textsuperscript{1,2}\thanks{Joint first authors} \hspace{0.3in} Karl Bladin\textsuperscript{1}\samethanks \hspace{0.3in} Yajie Zhao\textsuperscript{1}\samethanks \hspace{0.3in} Chinmay Chinara\textsuperscript{1} \hspace{0.3in} Owen Ingraham\textsuperscript{1} \hspace{0.3in}
Pengda Xiang\textsuperscript{1,2} \hspace{0.1in} Xinglei Ren\textsuperscript{1} \hspace{0.1in}
Pratusha Prasad\textsuperscript{1} \hspace{0.1in} Bipin Kishore\textsuperscript{1} \hspace{0.1in} Jun Xing\textsuperscript{1} \hspace{0.1in} Hao Li \textsuperscript{1,2,3} 
\vspace{0pt}
\\
\textsuperscript{1}{USC Institute for Creative Technologies} \hspace{0.3in} \textsuperscript{2}{University of Southern California} \hspace{0.3in} \textsuperscript{3}{Pinscreen}
}

\twocolumn[{%
\renewcommand\twocolumn[1][]{#1}%
\maketitle
\vspace{-0.3in}
\begin{center}
    \centering
   \vspace{-14pt}
  \includegraphics[width=\linewidth]{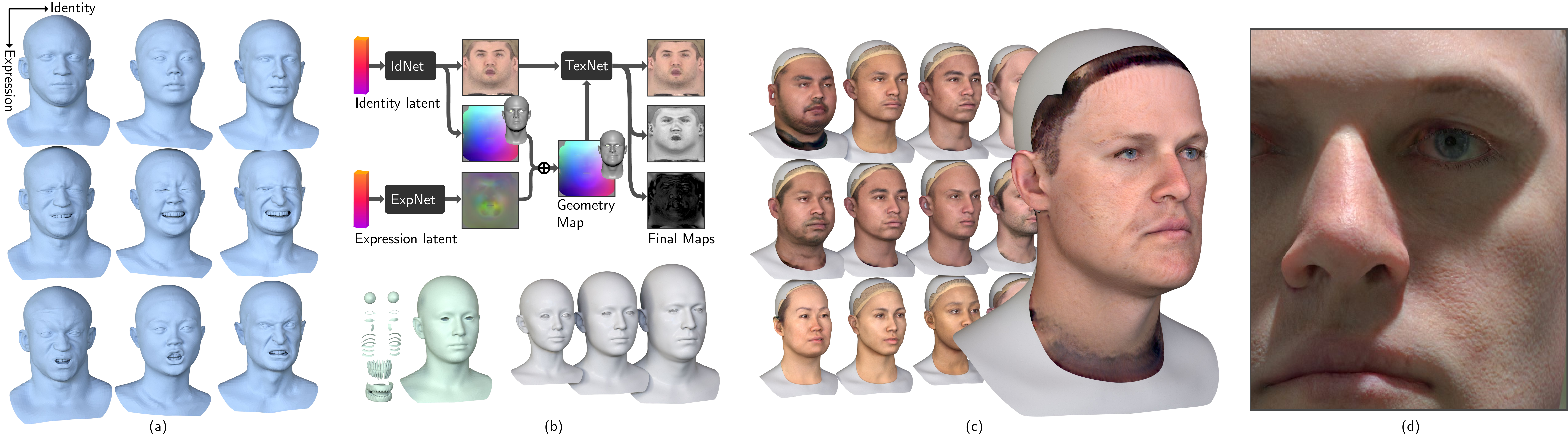}\\
  \vspace{10pt}
   \captionof{figure}{\small{We introduce a comprehensive framework for learning physically based face models from highly constrained facial scan data. Our deep learning based approach for 3D morphable face modeling seizes the fidelity of nearly 4000 high resolution face scans encompassing expression and identity separation (a). The model (b) combines a multitude of anatomical and physically based face attributes to generate an infinite number of digitized faces (c). Our model generates faces at pore level geometry resolution (d).}}
    \label{fig:teaser}
\end{center}
}]

\saythanks

\begin{abstract}
 \vspace{-9pt}
Based on a combined data set of 4000 high resolution facial scans, we introduce a non-linear morphable face model, capable of producing multifarious face geometry of pore-level resolution, coupled with material attributes for use in physically-based rendering.
We aim to maximize the variety of face identities, while increasing the robustness of correspondence between unique components, including middle-frequency geometry, albedo maps, specular intensity maps and high-frequency displacement details. Our deep learning based generative model learns to correlate albedo and geometry, which ensures the anatomical correctness of the generated assets. We demonstrate potential use of our generative model for novel identity generation, model fitting, interpolation, animation, high fidelity data visualization, and low-to-high resolution data domain transferring. We hope the release of this generative model will encourage further cooperation between all graphics, vision, and data focused professionals, while demonstrating the cumulative value of every individual's complete biometric profile. 
\end{abstract}

\section{Introduction}

Graphical virtual representations of humans are at the center of many endeavors in the fields of computer vision and graphics, with applications ranging from cultural media such as video games, film, and telecommunication to medical, biometric modeling, and forensics \cite{egger20193d}.

Designing, modeling, and acquiring high fidelity data for face models of virtual characters is costly and requires specialized scanning equipment and a team of skilled artists and engineers \cite{Ghosh2011MultiviewFC, Beeler2011HighqualityPF, Seol:2016:CAF:2947688.2947693}. Due to limiting and restrictive data policies of VFX studios, in conjunction with the absence of a shared platform that regards the sovereignty of, and incentives for the individuals’ data contributions, there is a large discrepancy in the fidelity of models trained on publicly available data, and those used in large budget game and film production.
A single, unified model would democratize the use of generated assets, shorten production cycles and boost quality and consistency, while incentivizing innovative applications in many markets and fields of research.

The unification of a facial scan data set in a \textit{3D morphable face model} (3DMM) \cite{blanz1999morphable,2014FaceWarehouseA3,Thies2018Face2FaceRF,egger20193d} promotes the favorable property of representing facial scan data in a compact form, retaining the statistical properties of the source without exposing the characteristics of any individual data point in the original data set.

Previous methods, including traditional methods \cite{blanz1999morphable, 2014FaceWarehouseA3, li2017learning, ploumpis2019combining,gerig2018morphable,booth20163d}, or deep learning \cite{tran2019towards,shamai2019synthesizing} to represent 3D face shapes; lack high resolution (sub-millimeter, $<1mm$) geometric detail, use limited representations of facial anatomy, or forgo the physically based material properties required by modern visual effects (VFX) production pipelines.
Physically based material intrinsics have proven difficult to estimate through the optimization of unconstrained image data due to ambiguities and local minima in analisys-by-synthesis problems, while highly constrained data capture remains percise but expensive \cite{egger20193d}. Although variations occur due to different applications, most face representations used in VFX employ a set of texture maps of at least ${4096 \times 4096 \: (4K)}$ pixels resolution. At a minimum, this set encorporates \textit{diffuse albedo}, \textit{specular intensity}, and \textit{displacement} (or \textit{surface normals}).

Our goal is to build a physically-based, high-resolution generative face model to begin bridging these parallel, but in some ways divergent, visualization fields; aligning the efforts of vision and graphics researchers. Building such a model requires high-resolution facial geometry, material capturing and automatic registration of multiple assets. The handling of said data has traditionally required extensive manual work, thus scaling such a database is non-trivial. For the model to be light weight these data need to be compressed into a compact form that enables controlled reconstruction based on novel input. Traditional methods such as PCA \cite{blanz1999morphable} and bi-linear models \cite{2014FaceWarehouseA3} $-$ which are limited by memory size, computing power, and smoothing due to inherent linearity $-$ are not suitable for high-resolution data.

By leveraging state-of-the-art physically-based facial scanning \cite{Ghosh2011MultiviewFC, legendre2018efficient}, in a \textit{Light Stage} setting, we enable acquisition of \textit{diffuse albedo} and \textit{specular intensity} texture maps in addition to $4K$ displacement.
All scans are registered using an automated pipeline that considers pose, geometry, anatomical morphometrics, and dense correspondence of $26$ expressions per subject.
A shared 2D UV parameterization data format \cite{gecer2019synthesizing, tran2019learning, shamai2019synthesizing}, enables training of a non-linear 3DMM, while the head, eyes, and teeth are represented using a linear PCA model. Hence, we propose a hybrid approach to enable a wide set of head geometry assets as well as avoiding the assumption of linearity in face deformations.

Our model fully disentangles identity from expressions, and provides manipulation using a pair of low dimensional feature vectors. To generate coupled geometry and albedo, we designed a joint discriminator to ensure consistency, along with two separate discriminators to maintain their individual quality. Inference and up-scaling of before-mentioned skin intrinsics enable recovery of $4K$ resolution texture maps.

Our main contributions are:
\begin{itemize}
    \item The first published upscaling of a database of high resolution ($4K$) physically based face model assets.
    
    \item A cascading generative face model, enabling control of identity and expressions, as well as physically based surface materials modeled in a low dimensional feature space.
    
    \item The first morphable face model built for full 3D real time \textit{and} offline rendering applications, with more relevant anatomical face parts than previously seen.
\end{itemize}

\section{Related Work}


\paragraph{Facial Capture Systems}

Physical object scanning devices span a wide range of categories; from single RGB cameras \cite{Garrido:2013:RDD,Shi2014AutomaticAO}, to active \cite{Alexander2009TheDE,Ghosh2011MultiviewFC}, and passive \cite{Beeler2010HighqualitySC} light stereo capture setups, and depth sensors based on time-of-flight or stereo re-projection. \textit{Multi-view stereo-photogrammetry} (MVS) \cite{Beeler2010HighqualitySC} is the most readily available method for 3D face capturing.
However, due to its many advantages over other methods (capture speed, physically-based material capturing, resolution), \textit{polarized spherical gradient illumination} scanning \cite{Ghosh2011MultiviewFC} remains state-of-the-art for high-resolution facial scanning.
A mesoscopic geometry reconstruction is bootstrapped using an MVS prior, utilizing omni-directional illumination, and progressively finalized using a process known as \textit{photometric stereo} \cite{Ghosh2011MultiviewFC}. The algorithm promotes the physical reflectance properties of dielectric materials such as skin; specifically the separable nature of specular and subsurface light reflections \cite{Ma2007RapidAO}. This enables accurate estimation of diffuse albedo and specular intensity as well as pore-level detailed geometry.

\nothing{
Facial scan registration is commonly performed using prediction of facial landmarks that are common for each scan subject. Although landmarks provide a sparse correspondence, they do not carry enough information to correspond different expressions of a single subject to pore level accuracy. Beeler \etal \cite{Beeler2011HighqualityPF} corresponded a sequence of multi view frames using image space optical flow. Fyffe \etal \cite{Fyffe2017MultiViewSO} showed that reconstruction and registration of a performance capture sequence can be performed by posing a joint optimization problem that can be solved using optical flow with a volumetric Laplacian constraint.
}

\paragraph{3D Morphable Face Models}
The first published work on morphable face models by Blanz and Vetter \cite{blanz1999morphable} represented faces as dense surface geometry and texture, and modeled both variations as separate PCA models learned from around 200 subject scans. To allow intuitive control; attributes, such as gender and fullness of faces, were mapped to components of the PCA parameter space. This model, known as the \textit{Basel Face Model} \cite{Paysan2009A3F} was released for use in the research community, and was later extended to a more diverse linear face model learnt from around 10,000 scans \cite{booth20163d,Booth2017LargeS3}. 

To incorporate facial expressions, Vlasic \etal \cite{Vlasic2005FaceTW} proposed a multi-linear model to jointly estimate the variations in identity, viseme, and expression, and Cao \etal \cite{2014FaceWarehouseA3} built a comprehensive bi-linear model (identity and expression) covering 20 different expressions from 150 subjects learned from RGBD data. Both of these models adopt a tensor-based method under the assumption that facial expressions can be modeled using a small number of discrete poses, corresponded between subjects. More recently, Li \etal. \cite{li2017learning} released the FLAME model, which incorporates both pose-dependent corrective blendshapes, and additional global identity and expression blendshapes learnt from a large number of 4D scans.

To enable adaptive, high level, semantic control over face deformations, various locality-based face models have been proposed. Neumann \etal \cite{Neumann2013SparseLD} extract sparse and spatially localized deformation modes, and Brunton \etal \cite{Brunton2014MultilinearWA} use a large number of localized multilinear wavelet modes.
As a framework for anatomically accurate local face deformations, the \textit{Facial Action Coding System} (FACS) by Ekman \cite{Ekman1978FacialAC} is widely adopted.
It decomposes facial movements into basic action units attributed to the full range of motion of all facial muscles.

Morphable face models have been widely used for applications like face fitting \cite{blanz1999morphable}, expression manipulation \cite{2014FaceWarehouseA3}, real-time tracking \cite{Thies2018Face2FaceRF}, as well as in products like Apple's ARKit. However, their use cases are often limited by the resolution of the source data and restrictions of linear models causing smoothing in middle and high frequency geometry details (e.g. wrinkles, and pores). Moreover, to the best of our knowledge, all existing morphable face models generate texture and geometry separately, without considering the correlation between them. 
Given the specific and varied ways in which age, gender, and ethnicity are manifested within the spectrum of human life, ignoring such correlation will cause artifacts; e.g. pairing an African-influenced albedo to an Asian-influenced geometry.

\begin{figure}[t]
\begin{center}
\includegraphics[width=0.4\textwidth]{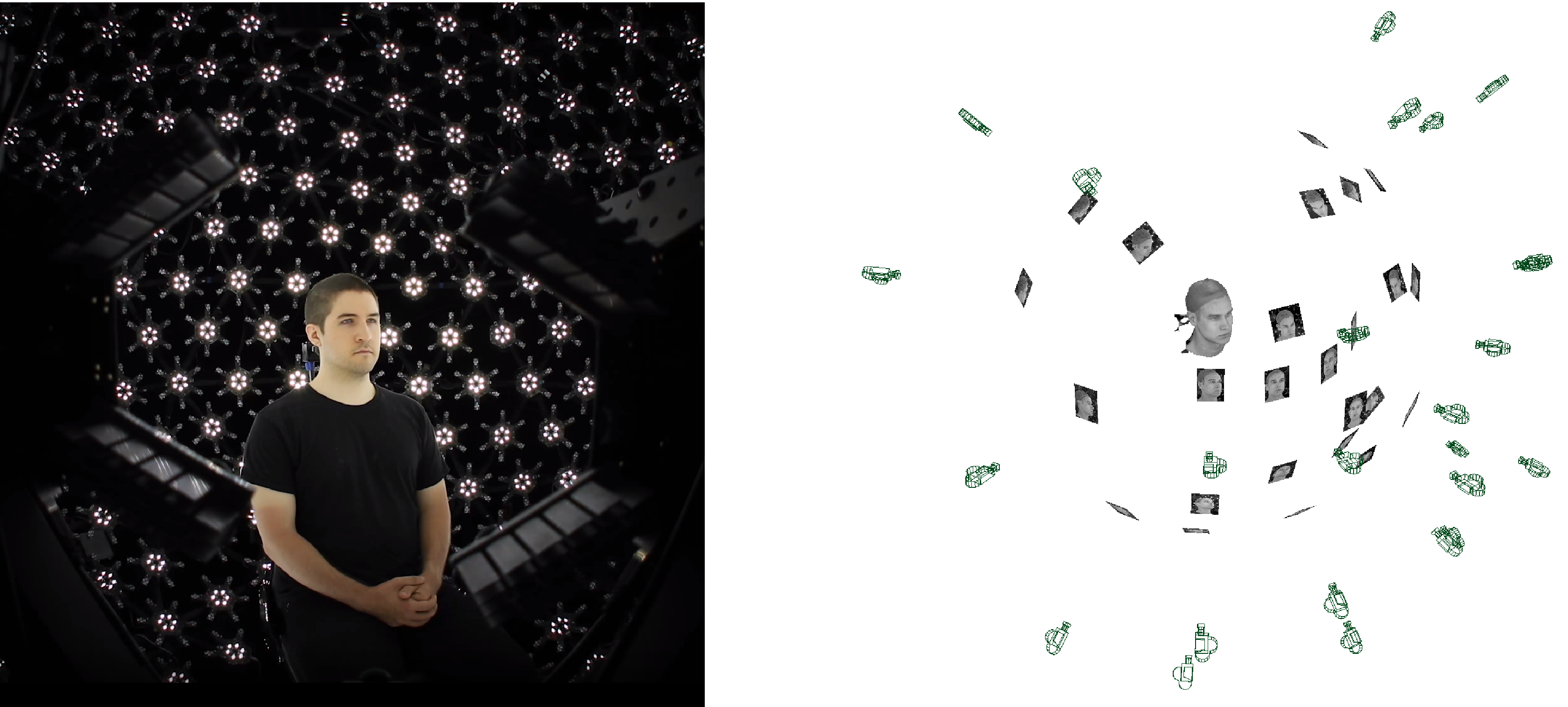}
\end{center}
\vspace{10pt}
\caption{Capture system and camera setup. Left: Light Stage capturing system.  Right: camera layout.}
\label{fig:light-stage}
\end{figure}

\paragraph{Image-based Detail Inference} To augment the quality of existing 3DMMs, many works have been proposed to infer the fine-level details from image data. Skin detail can be synthesized using data-driven texture synthesis \cite{Haro2001RealtimePP} or statistical skin detail models \cite{Golovinskiy2006ASM}. Cao \etal \cite{Cao:2015:RHF} used a probability map to locally regress the medium-scale geometry details, where a regressor was trained from captured patch pairs of high-resolution geometry and appearance. Saito \etal \cite{Saito2016PhotorealisticFT} presented a texture inference technique using a deep neural network-based feature correlation analysis.

GAN-based Image-to-Image frameworks \cite{Isola2016ImagetoImageTW} have proven to be powerful for high-quality detail synthesis, such as the coarse \cite{Trigeorgis2017FaceN}, medium \cite{Sela2017UnrestrictedFG} or even mesoscopic \cite{Huynh2018MesoscopicFG} scale facial geometry inferred directly from images. Beside geometry, Yamaguchi \etal \cite{Yamaguchi2018HighfidelityFR} presented a comprehensive method to infer facial reflectance maps (diffuse albedo, specular intensity, and medium- and high-frequency displacement) based on single image inputs.
More recently, Nagano \etal \cite{Nagano2018paGANRA} proposed a framework for synthesizing arbitrary expressions both in image space and UV texture space, from a single portrait image.
Although these methods can synthesize facial geometry or/and texture maps from a given image, they don't provide explicit parametric controls of the generated result.

\section{Database}
\label{sec:dataPreparation}
\subsection{Data Capturing and Processing}
\label{sec:data-capturing}

\paragraph{Data Capturing}
Our \emph{Light Stage} scan system employs \textit{photometric stereo} \cite{Ghosh2011MultiviewFC} in combination with monochrome color reconstruction using \textit{polarization promotion} \cite{legendre2018efficient} to allow for pore level accuracy in both the geometry reconstruction and the reflectance maps. The camera setup (Fig.\ref{fig:light-stage}) was designed for rapid, database scale, acquisition by the use of Ximea machine vision cameras which enable faster streaming and wider depth of field than traditional \textit{DSLRs} \cite{legendre2018efficient}. The total set of 25 cameras consists of eight 12MP \nothing{model \textit{MC124MG-SY-UB} 12MP} monochrome cameras, eight \nothing{model \textit{MC124CG-SY-UB} }12MP color cameras, and nine \nothing{model \textit{MQ042MG-CM} }4MP monochrome cameras. The 12MP monochrome cameras allow for pore level geometry, albedo, and specular reflectance reconstruction, while the additional cameras aid in stereo base mesh-prior reconstruction. 

\begin{figure}[t]
\begin{center}
\begin{subfigure}[b]{0.24\textwidth}
\includegraphics[width=1\textwidth]{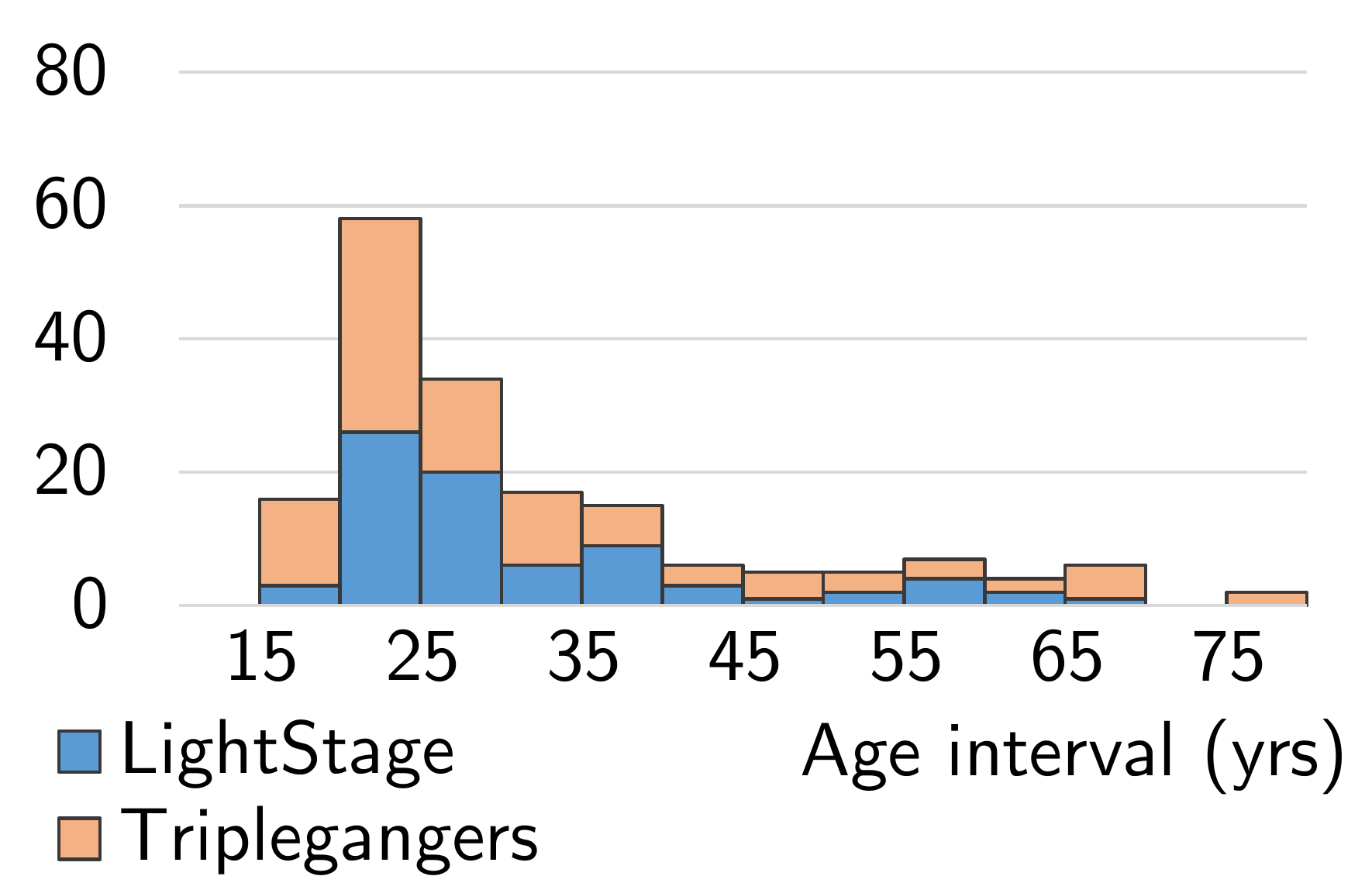}
\caption{Age distribution}
\end{subfigure}%
\begin{subfigure}[b]{0.24\textwidth}
\includegraphics[width=1\textwidth]{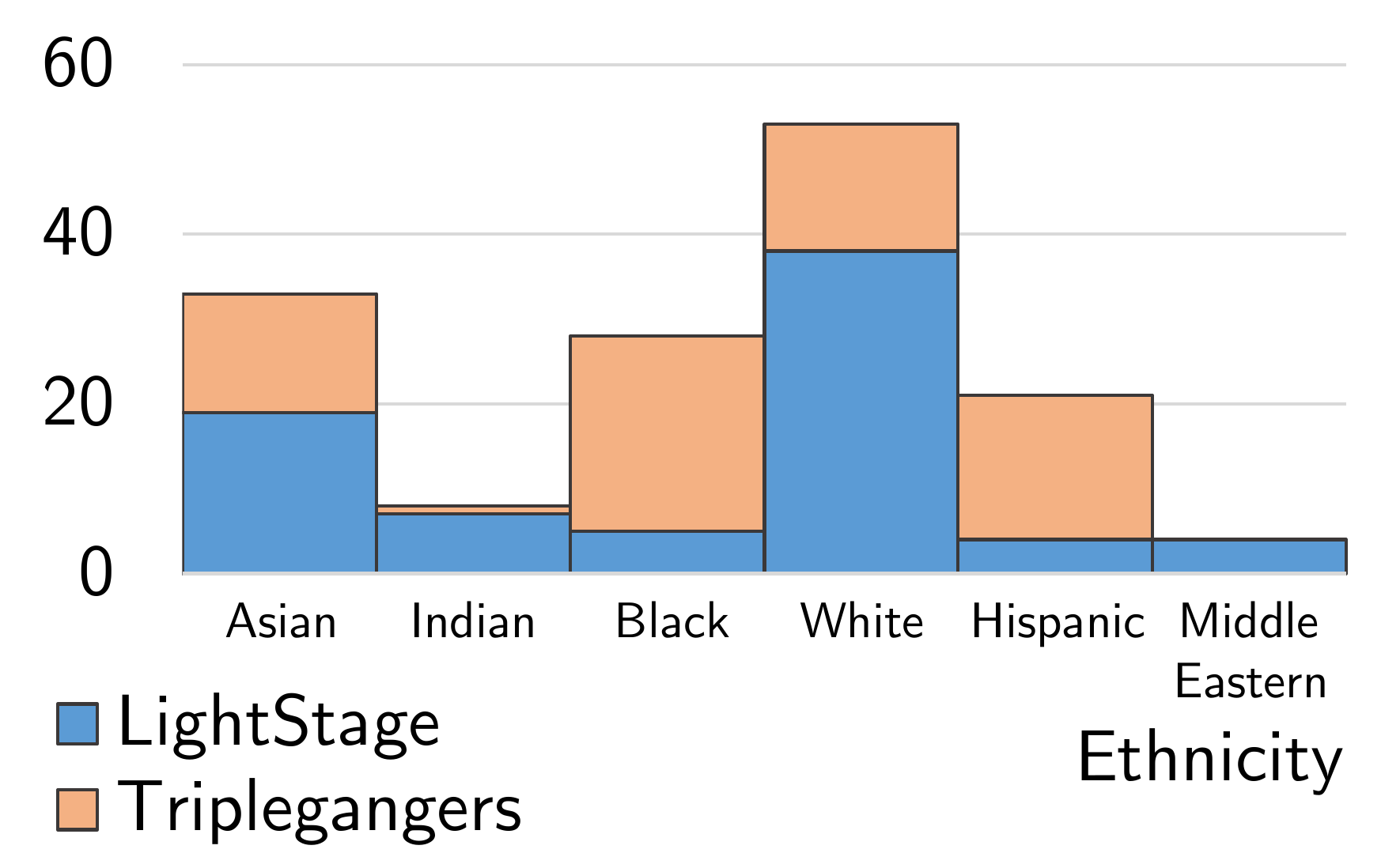}
\caption{Ethnicity distribution}
\end{subfigure}%
\end{center}
\caption{Distribution of age (a) and ethnicity (b) in the data sets.}
\label{fig:AgeEthnicity}
\end{figure}

To capture consistent data across multiple subjects with maximized expressiveness, we devised a FACS set \cite{Ekman1978FacialAC} which combines 40 \textit{action units} to a condensed set of 26 expressions. In total, 79 subjects, 34 female, and 45 male, ranging from age 18 to 67, were scanned performing the 26 expressions.
To increase diversity, we combined the data set with a selection of 99 \emph{Triplegangers} \cite{triplegangers} full head scans; each with 20 expressions.
Resolution and extent of the two data sets are shown in Table \ref{tab:datasets_statistics}. Fig. \ref{fig:AgeEthnicity} shows the age and ethnicity (multiple choice) distributions of the source data.

\begin{table}[t]
  \begin{center}
    \begin{tabular}{l|c|c|c|c|r} 
    \toprule[2pt]
       &\textbf{(a)} & \textbf{(b)} & \textbf{(c)} & \textbf{(d)} & \textbf{(e)} \\
      \hline\hline
      LS & $4k \times 4k$ & $3.9M$ & $4k \times 4k$ & 79 & 26\\
      TG & $8k \times 8k$ & $3.5M$ & N/A & 99 & 20\\
    \bottomrule[2pt]
    \end{tabular}
    \vspace{10pt}
    \caption{\small{Resolution and extent of the datasets. (a). Albedo resolution. (b). Geometry resolution. (c). Specular intensity resolution. (d) $\#$ of subjects. (f). $\#$ of expressions per subject. }}
    \label{tab:datasets_statistics}
  \end{center}
\end{table}

\paragraph{Processing Pipeline.}
Starting from the multi-view imagery, a neutral scan base mesh is reconstructed using MVS. Then a linear PCA model in our topology (See Fig.\ref{fig:3dmm}) based on a combination and extrapolation of two existing models (Basel \cite{Paysan2009A3F} and Face Warehouse \cite{2014FaceWarehouseA3}) is used to fit the mesh. Next, Laplacian deformation is applied to deform the face area to further minimize the surface-to-surface error. Cases of inaccurate fitting were manually modeled and fitted to retain the fitting accuracy of the eyeballs, mouth sockets and skull shapes. The resulting set of neutral scans were immediately added to the PCA basis for registering new scans.
We fit expressions using generic blendshapes and non-rigid ICP \cite{Li:2008:GCO}. Additionally, to retain texture space and surface correspondence, image space optical flow from neutral to expression scan is added from 13 different virtual camera views as additional dense constraint in the final Laplacian deformation of the face surface.

\subsection{Training Data Preparation}
\label{sec:2d-geo}

\begin{figure}[t]
\begin{center}
\includegraphics[width=0.45\textwidth]{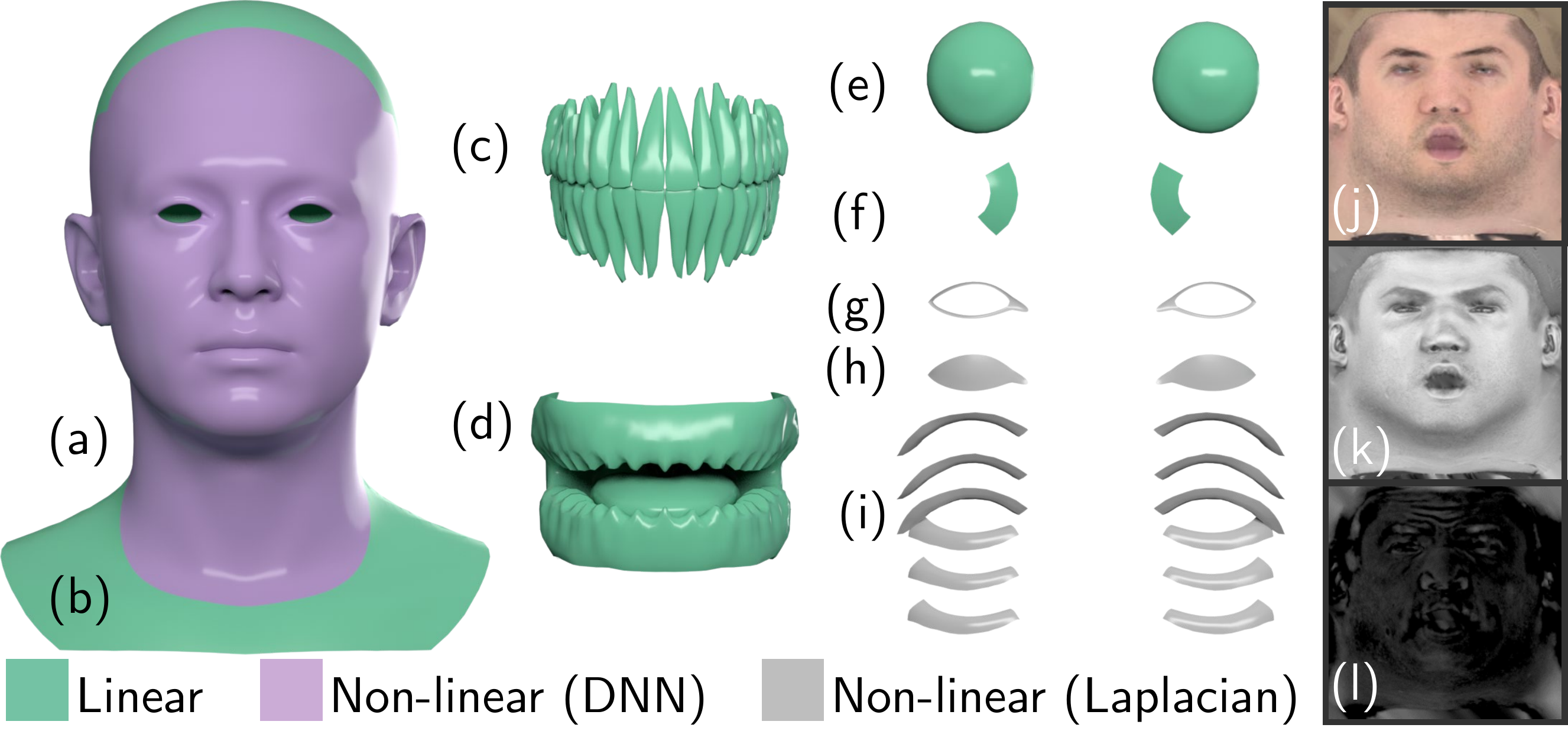}
\end{center}
   \caption{Our generic face model consists of multiple geometries constrained by different types of deformation. In addition to face (a), head, and neck (b), our model represents teeth (c), gums (d), eyeballs (e), eye blending (f), lacrimal fluid (g), eye occlusion (h), and eyelashes (i). Texture maps provide high resolution (4K) albedo (j), specularity (k), and geometry through displacement (l).
   }
\label{fig:3dmm}
\end{figure}

\begin{figure}[t]
\begin{center}
\includegraphics[width=0.45\textwidth]{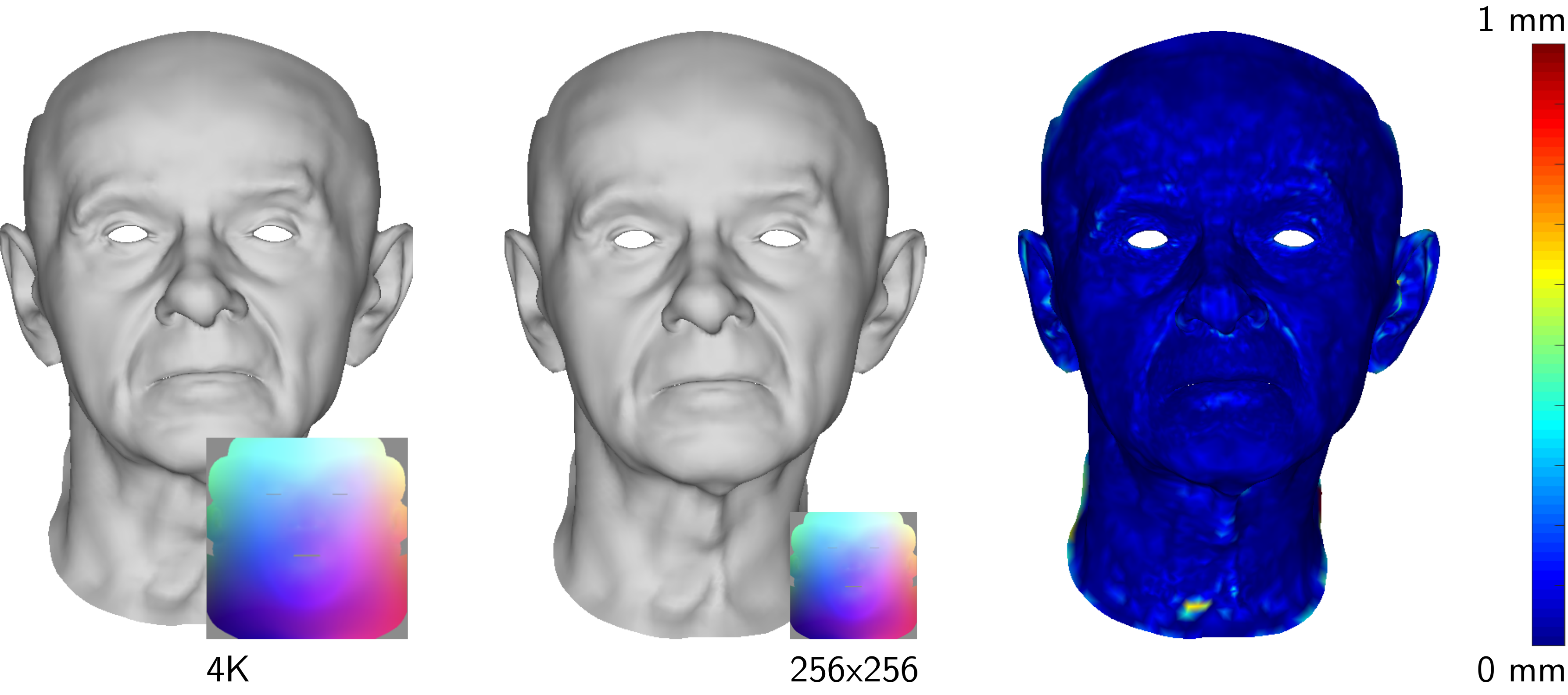}
\end{center}
   \caption{Comparison of base mesh geometry resolutions. Left: Base geometry reconstructed in $4K$ resolution. Middle: Base geometry reconstructed in $256 \times 256$ resolution. Right: Error map showing the Hausdorff distance in the range $(0 mm, 1 mm)$, with a mean error of $0.068 mm$.}
\label{fig:prove256}
\end{figure}

\paragraph{Data format.}
The full set of the generic model consists of a hybrid geometry and texture maps (\textit{albedo}, \textit{specular intensity}, and \textit{displacement}) encoded in $4K$ resolution, as illustrated in Fig. \ref{fig:3dmm}.
To enable joint learning of the correlation between geometry and albedo, 3D vertex positions are rasterized to a three channel HDR bitmap of $256 \times 256$ pixels resolution.
The face area (pink in Fig. \ref{fig:3dmm}) used to learn the geometry distribution in our non-linear generative model consists of $m = 11892$ vertices, which, if evenly spread out in texture space, would require a bitmap of resolution greater or equal to $\ceil{\sqrt{2 \times m}}^2 = 155 \times 155$, according to Nyquist's resampling theorem.
As shown in Fig. \ref{fig:prove256}, the proposed resolution is enough to recover middle-frequency detail. This relatively low resolution base geometry representation enables great simplification in training data load.

\begin{figure*}
\begin{center}
  \includegraphics[width=1\linewidth]{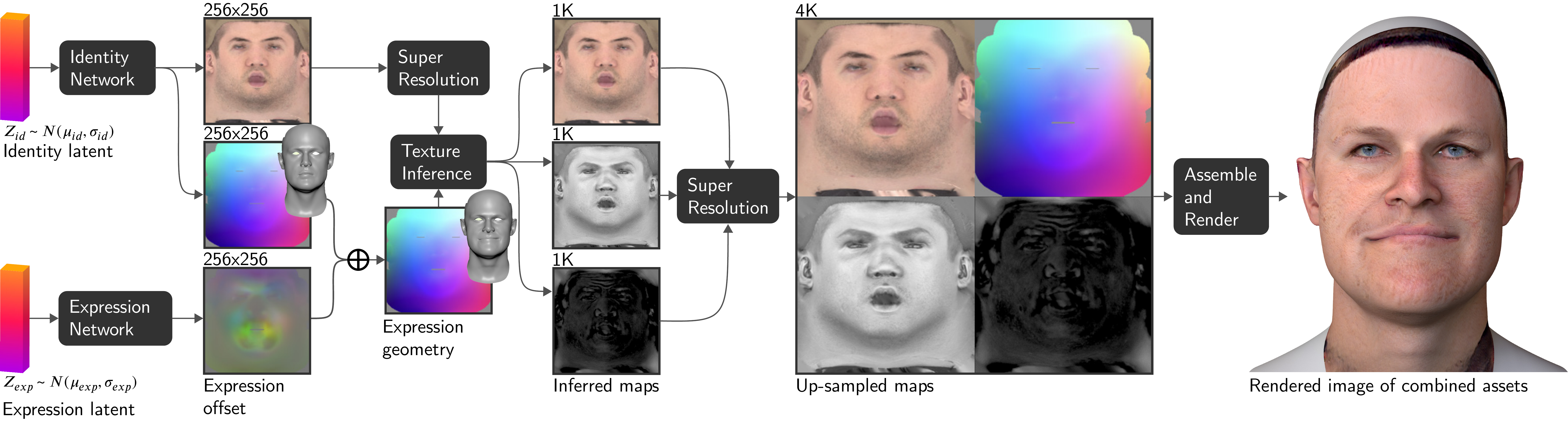}
  \end{center}
   \caption{Overview of generative pipeline. Latent vectors for identity and expression serve as input for generating the final face model.}
\label{fig:overview}
\end{figure*}

\paragraph{Data Augmentation}
\label{sec:training-data}
Since the number of subjects is limited to 178 individuals, we apply two strategies to augment the data for identity training: 1) For each source albedo, we randomly sample a target albedo within the same ethnicity and gender in the data set using \cite{zhao2019mask} to transfer skin tones of target albedos to source albedos (these samples are restricted to datapoints of the same ethnicity), followed by an image enhancement~\cite{hacohen2011non} to improve the overall quality and remove artifacts. 2). For each neutral geometry, we add a very small expression offset using FaceWarehouse expression components with a small random weights($< \pm 0.5$ std) to loosen the constraints of ``neutral".  To augment the expressions, we add random expression offsets to generate fully controlled expressions.

\section{Generative Model}
\label{sec:pipeline}
An overview of our system is illustrated in Fig.~\ref{fig:overview}. 
Given a sampled latent code $Z_{id} \sim N(\mu_{id}, \sigma_{id})$, our \textit{Identity} network generates a consistent albedo and geometry pair of neutral expression.
We train an \textit{Expression} network to generate the expression offset that can be added to the neutral geometry.
We use random blendshape weights $Z_{exp} \sim N(\mu_{exp}, \sigma_{exp})$ as the expression network's input to enable manipulation of target semantic expressions.
We upscale the albedo and geometry maps to $1K$, and feed them into a transfer network \cite{wang2018pix2pixHD} to synthesize the corresponding $1K$ specular and displacement maps. Finally, all the maps except for the middle frequency geometry map are upscaled to $4K$ using \textit{Super-resolution} \cite{ledig2017photo}, as we observed that $256\times256$ pixels are sufficient to represent the details of the base geometry (Section~\ref{sec:2d-geo}). The details of each component are elaborated on in Section~\ref{sec:id_GAN}, \ref{sec:Exp_GAN}, and \ref{sec:inference}.

\subsection{Identity Network}
\label{sec:id_GAN}

\begin{figure}[t]
\begin{center}
\includegraphics[width=0.4\textwidth]{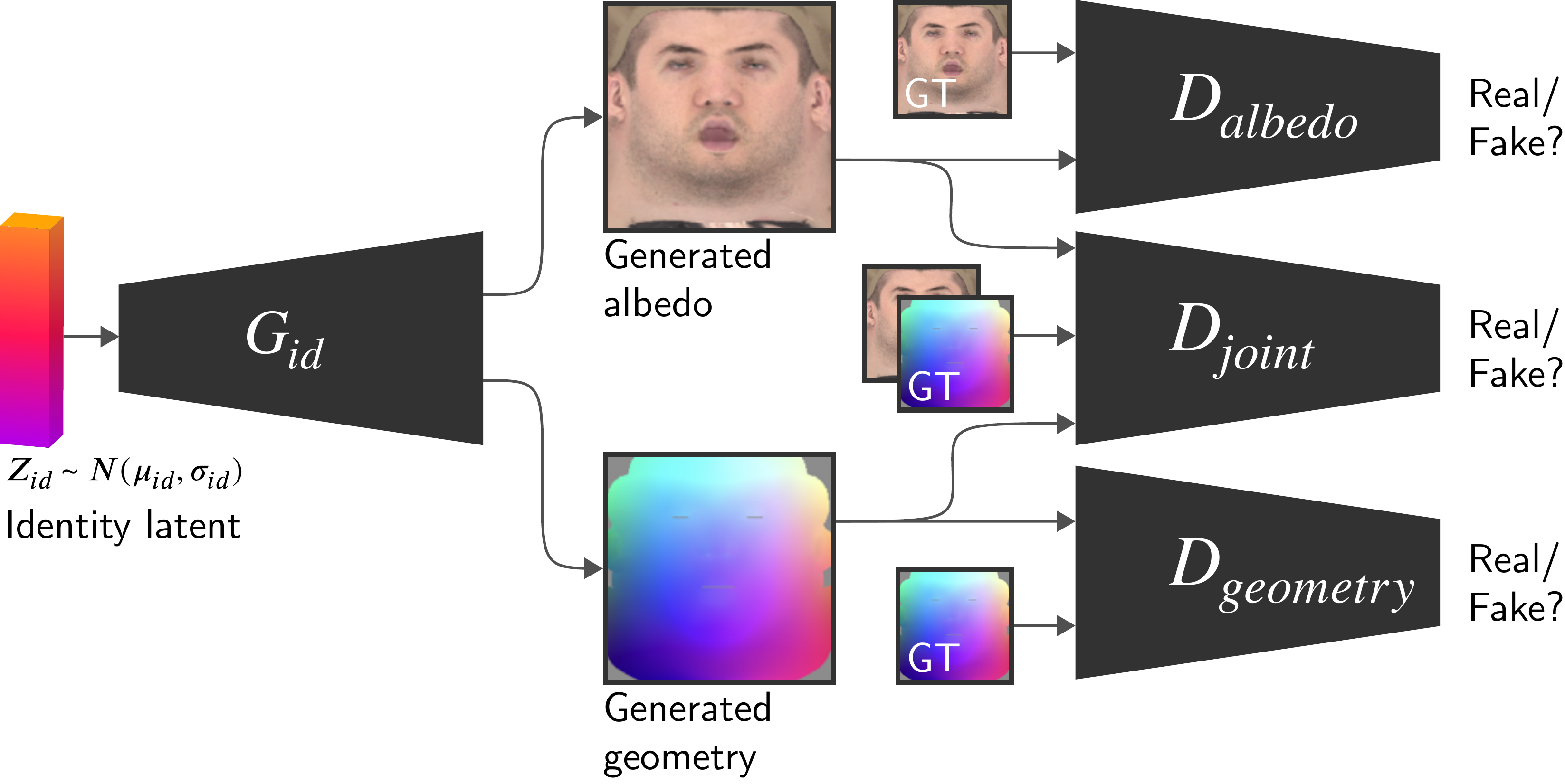}
\end{center}
\caption{\textit{Identity} generative network. The identity generator $G_{id}$ produces albedo and geometry which get checked against ground truth (GT) data by the discriminators, $D_{albedo}$, $D_{joint}$, and $D_{geometry}$ during training.}
\label{fig:Idnet}
\end{figure}

The goal of our \emph{Identity} network is to model the cross correlation between geometry and albedo to generate consistent, diverse and biologically accurate identities.
The network is built upon the \emph{Style-GAN} architecture~\cite{karras2019style}, that can produce high-quality, style-controllable sample images.

To achieve consistency, we designed 3 discriminators as shown in Fig.\ref{fig:Idnet}, including individual discriminators for albedo ($D_{albedo}$) and geometry ($D_{geometry}$), to ensure the quality and sharpness of the generated maps, and an additional joint discriminator ($D_{joint}$) to learn their correlated distribution. $D_{joint}$ is formulated as follows:
\begin{equation}
\begin{split}
\mathcal{L}_{\mathrm{adv}} & = \min _{G_{id}} \max _{D_{joint}}\
\mathbb{E}_{\boldsymbol{x}
	\sim
	p_\mathrm{data}(\boldsymbol{x})} \big[ \mathrm{log}\
D_{joint}(\boldsymbol{\mathcal{A}}) \big] + \\ &\mathbb{E}_{\boldsymbol{z} \sim
	p_{\boldsymbol{z}}(\boldsymbol{z})}\big[ \mathrm{log}\ (1 -
D_{joint}(G_{id}(\boldsymbol{z})))\big].
\end{split}
\end{equation}
where $p_\mathrm{data}(\boldsymbol{x})$ and $p_{\boldsymbol{z}}(\boldsymbol{z})$ represent the distributions of real paired albedo and geometry $\boldsymbol{x}$ and noise variables $\boldsymbol{z}$ in the domain of $\boldsymbol{\mathcal{A}}$ respectively.

\subsection{Expression Network}
\label{sec:Exp_GAN}

\begin{figure}[t]
\begin{center}
\includegraphics[width=0.4\textwidth]{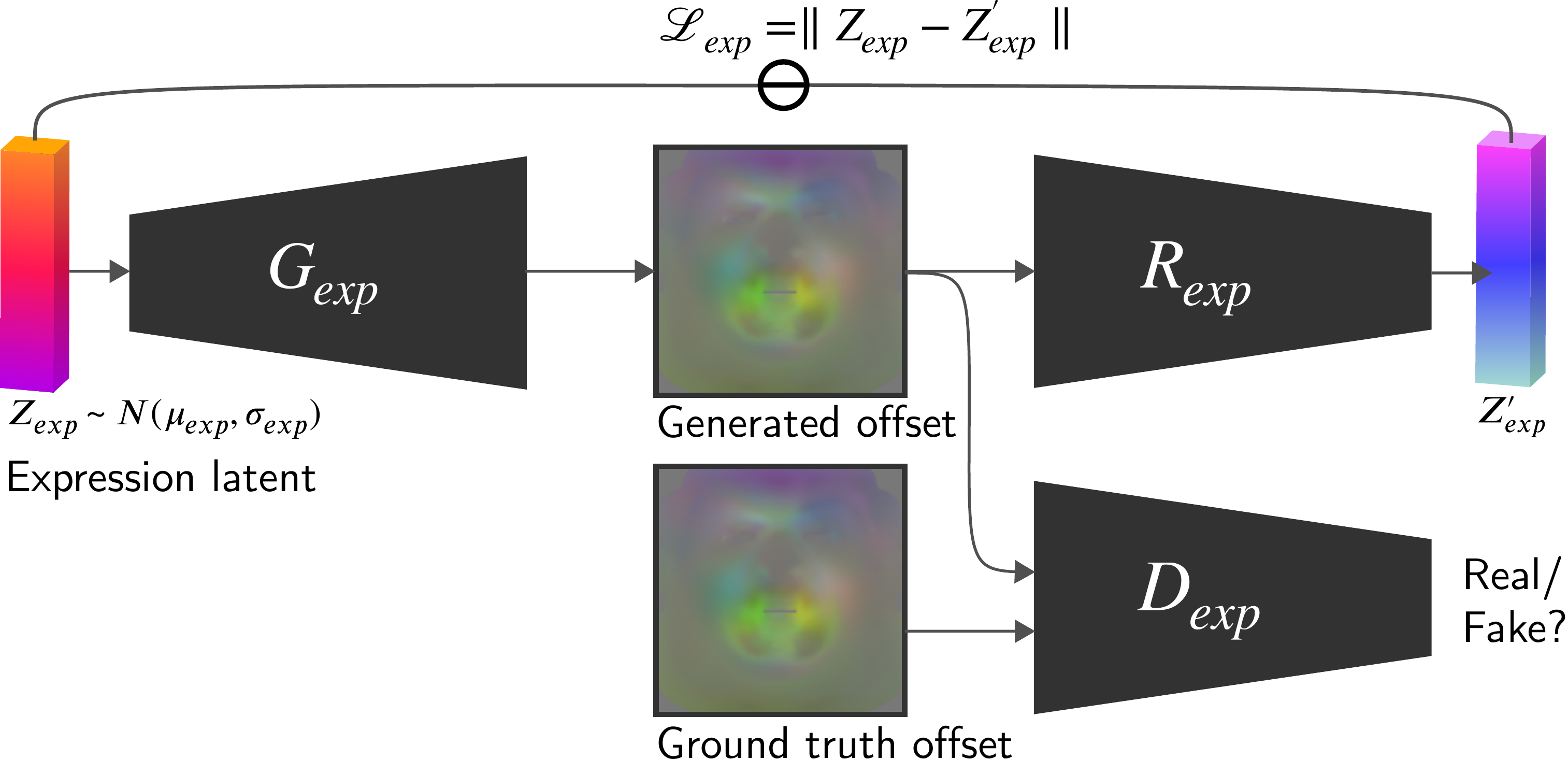}
\end{center}
\caption{\textit{Expression} generative network. The expression generator $G_{exp}$ generates offsets which get checked against ground truth offsets by the discriminator $D_{exp}$. The regressor $R_{exp}$ produces an estimate of the latent code $Z^{'}_{exp}$ so that the L1 loss $\mathcal{L}_{exp}$ can be modeled.}
\label{fig:expnet}
\end{figure}

To simplify the learning of a wide range of diverse expressions, we represent them using vector offset maps, which also makes the learning of expressions independent from identity.
Similar to the \emph{Identity} network, the expression network adopts \emph{Style-GAN} as the base structure.
To allow for intuitive control over expressions, we use the \textit{blendshape} weights, which correspond to the strength of 25 orthogonal facial activation units, as network input.
We introduce a pre-trained expression regression network $R_{exp}$ to predict the expression weights from the generated image, and force this prediction to be similar to the input latent code $Z_{exp}$. We then force the generator to understand the input latent code $Z_{exp}$ under the perspective of the pre-trained expression regression network. As a result, each dimension of the latent code $Z_{exp}$ will control the corresponding expression defined in the original blendshape set. The loss we introduce here is:

\begin{equation}
\mathcal{L}_{exp} =  \parallel Z_{exp} - Z^{'}_{exp} \parallel
\end{equation}
This loss, $\mathcal{L}_{exp}$, will be back propagated during training to enforce the orthogonality of each blending unit. We minimize the following losses to train the network:
\begin{equation}
\mathcal{L} = \mathcal{L}^{exp}_{l_2} + \beta_{1} \mathcal{L}^{exp}_{adv} + \beta_{2} \mathcal{L}_{exp}
\end{equation}
where $\mathcal{L}^{exp}_{l_2}$ is the $L_2$ reconstruction loss of the offset map and $\mathcal{L}^{exp}_{adv}$ is the discriminator loss. 

\subsection{Inference and Super-resolution }
\label{sec:inference}
Similar to~\cite{Yamaguchi2018HighfidelityFR}; upon obtaining  albedo and geometry maps (${256 \times 256}$), we use them to infer specular and displacement maps in ${1K}$ resolution.
In contrast to \cite{Yamaguchi2018HighfidelityFR}, using only albedo as input, we introduce the geometry map to form stronger constraints. For displacement, we adopted the method of~\cite{Yamaguchi2018HighfidelityFR,Huynh2018MesoscopicFG} to separate displacement in to individual high-frequency and low-frequency components, which makes the problem more tractable.
Before feeding the two inputs into the inference network \cite{wang2018pix2pixHD}, we up-sample the albedo to $1K$ using a super-resolution network similar to~\cite{ledig2017photo}. The geometry map is super-sampled using bi-linear interpolation.
The maps are further up-scaled from $1K$ to $4K$ using the same super-resolution network structure. Our method can be regarded as a two step cascading up-sampling strategy ($256$ to $1K$, and $1K$ to $4K$). This makes the training faster, and enables higher resolution in the final results.

\section{Implementation Details}

Our framework is implemented using Pytorch and all our networks are trained using two NVIDIA Quadro GV100s.
We follow the basic training schedule of Style-GAN \cite{karras2019style} with several modifications applied to the \textit{Expression} network, like by-passing the progressive training strategy as expression offsets are only distinguishable on relatively high resolution maps.
We also remove the noise injection layer, due to the input latent code $Z_{exp}$ which enables full control of the generated results. The regression module ($R_{exp}$-block in Fig.\ref{fig:expnet}) has the same structure as the discriminator $D_{exp}$, except for the number of channels in the last layer, as it serves as a discriminator during training.
The regression module is initially trained using synthetic unit expression data generated with neutral expression and $Face Warehouse$ expression components, and then fine-tuned on scanned expression data. During training, $R_{exp}$, is fixed without updating parameters.
The \textit{Expression} network is trained with a constant batch size of 128 on 256x256-pixel images for 40 hours. The \textit{Identity} network is trained by progressively reducing the batch size from 1024 to 128 on growing image sizes ranging from 8x8 to 256x256 pixels, for 80 hours.

\section{Experiments And Evaluations}
\subsection{Results}
In Fig.\ref{fig:quality}, we show the quality of our generated model rendered using Arnold. The direct output of our generative model provides all the assets necessary for physically-based rendering in software such as Maya, Unreal Engine, or Unity 3D. We also show the effect of each generated component.

\begin{figure}[t]
\begin{center}
\includegraphics[width=0.45\textwidth]{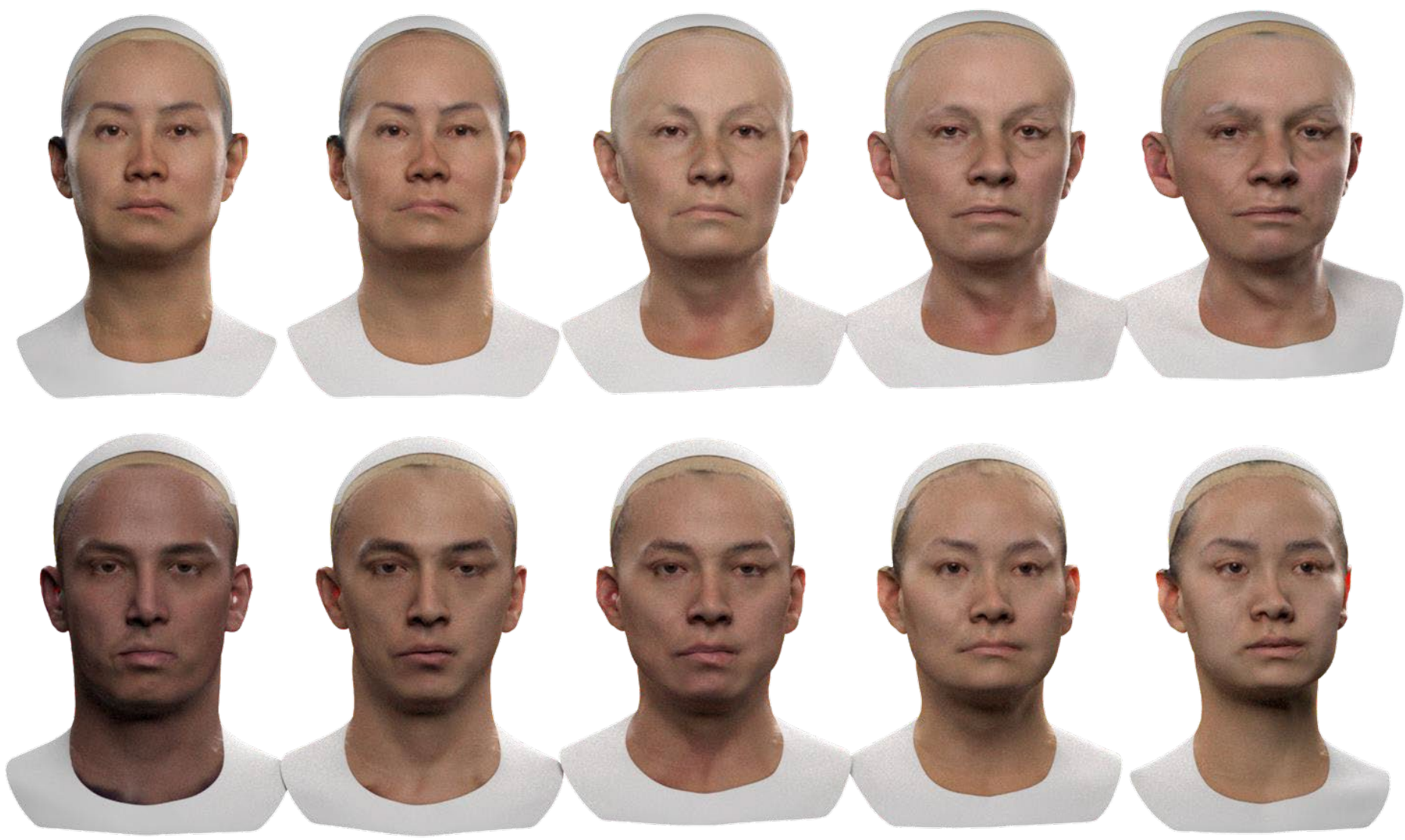}
\vspace{5pt}
\end{center}
   \caption{Non-linear identity interpolation between generated subjects. Age (top) and gender (bottom) are interpolated from left to right.}
\vspace{5pt}
\label{fig:ageInterploation}
\end{figure}

\begin{figure}[t]
\begin{center}
\includegraphics[width=0.47\textwidth]{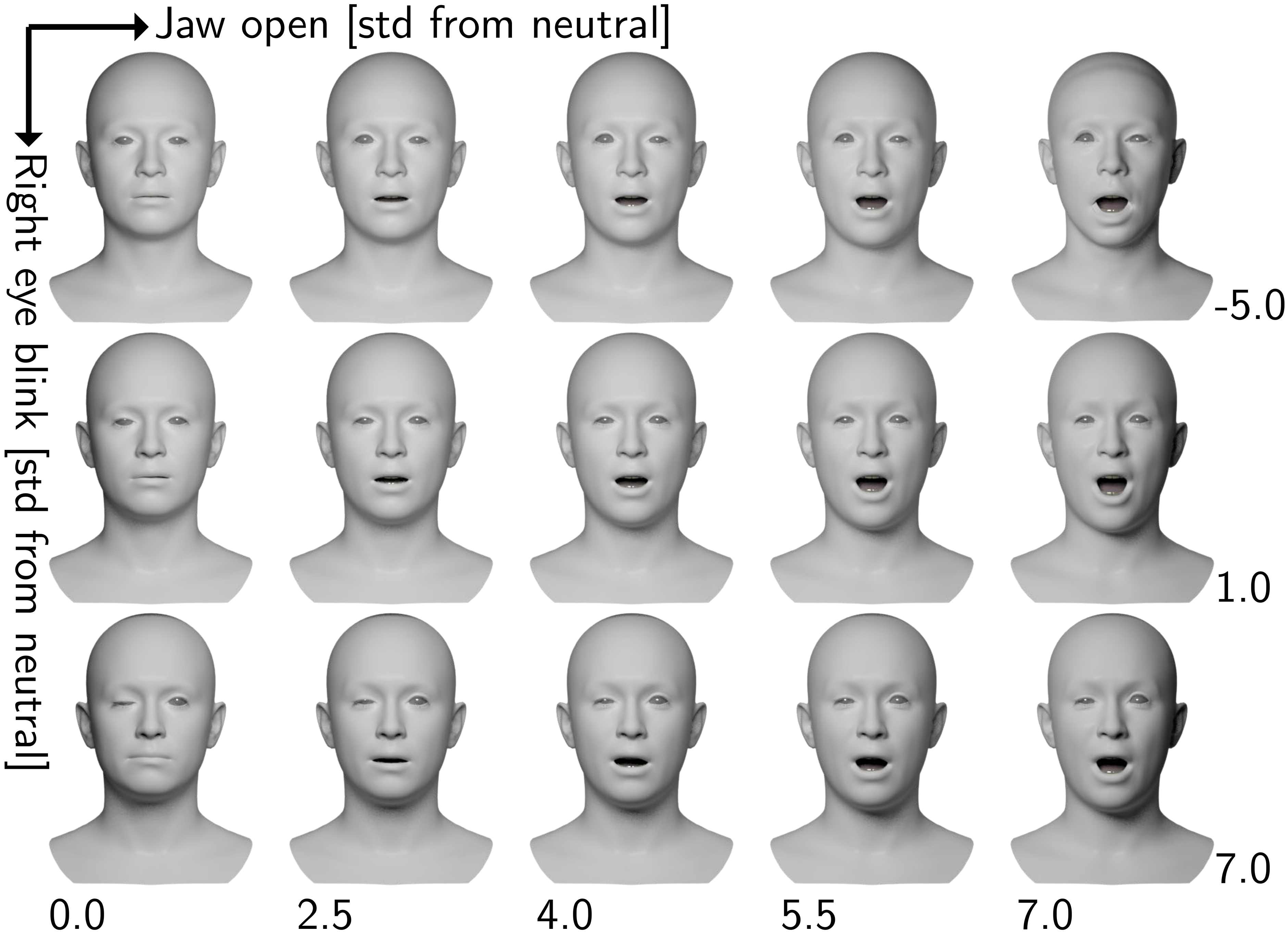}
\end{center}
\vspace{5pt}
\caption{Non linear expression interpolation using generative expression network. Combinations of two example shapes are displayed in a grid where the number of standard deviations from the generic neutral model define the extent of an expression shape.}
\label{fig:Expression Control}
\end{figure}

\subsection{Qualitative Evaluation}
We show identity interpolation in Fig.\ref{fig:ageInterploation}. The interpolation in latent space reflects both albedo and geometry. In contrast to linear blending, our interpolation generates subjects belonging to a natural statistical distribution.

In Fig.\ref{fig:Expression Control}, we show the generation and interpolation of our non-linear expression model. We pick two orthogonal blendshapes for each axis and gradually change the input weights. Smooth interpolation in vector space will lead to a smooth interpolation in model space.

We show nearest neighbors for generated models in the training set in Fig.\ref{fig:nn_search}. These are found based on point-wise Euclidean distance in geometry. Albedos are compared to prove our ability to generate new models that are not merely recreations of the training set.

\begin{figure*}[h!]
\begin{center}
\begin{subfigure}[t]{0.16\textwidth}
\includegraphics[width=1\textwidth]{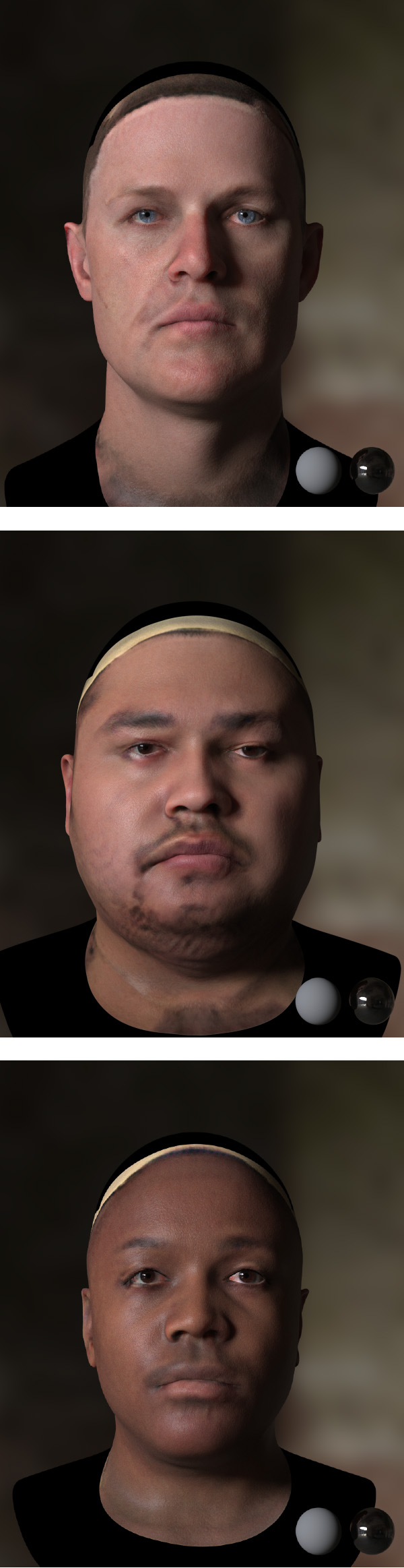}
\caption{}
\end{subfigure}
\begin{subfigure}[t]{0.16\textwidth}
\includegraphics[width=1\textwidth]{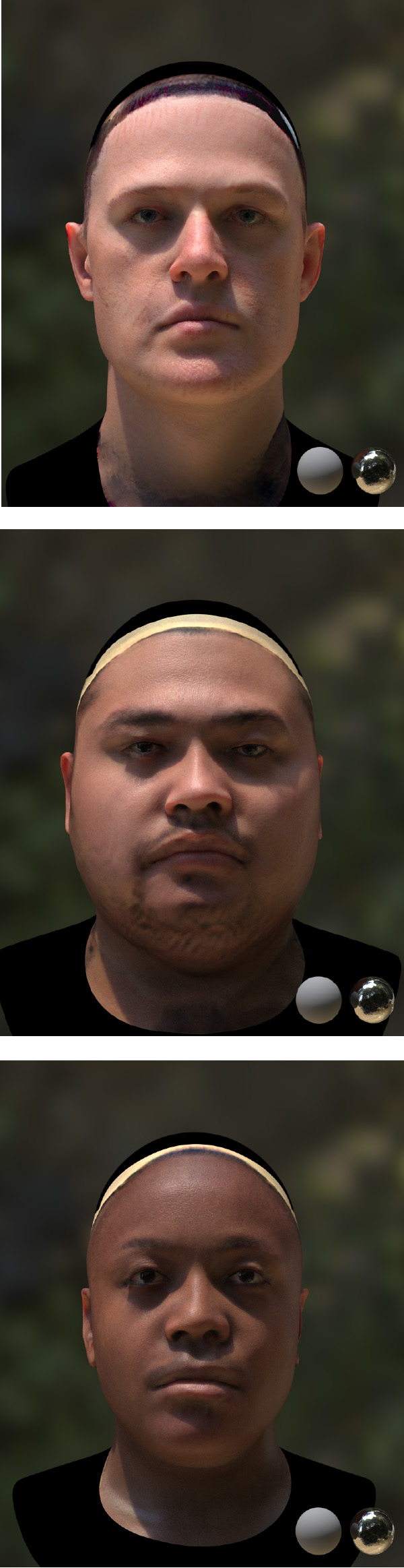}
\caption{}
\end{subfigure}
\begin{subfigure}[t]{0.16\textwidth}
\includegraphics[width=1\textwidth]{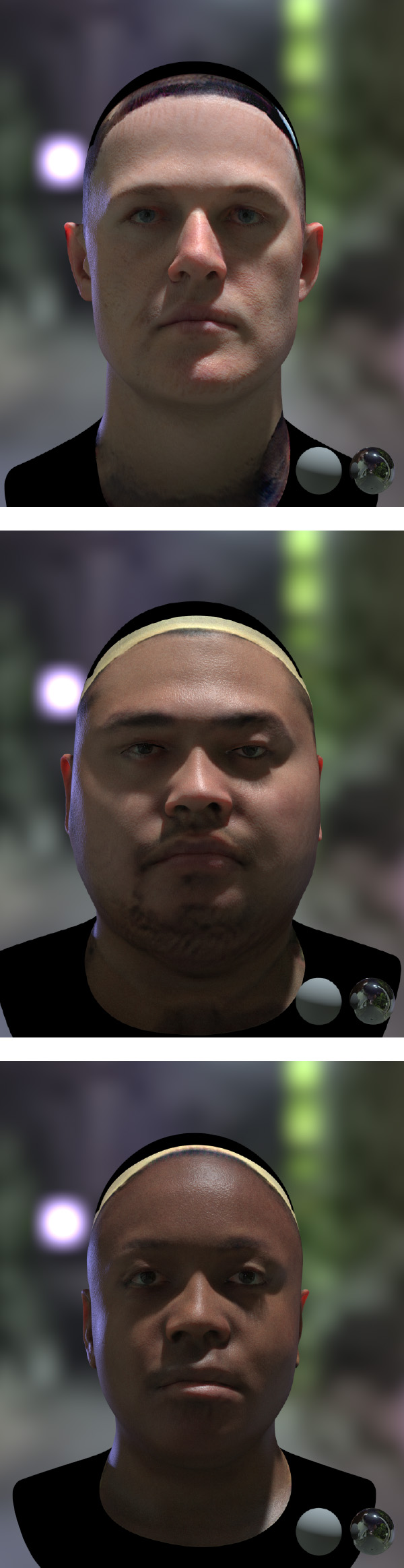}
\caption{}
\end{subfigure}
\begin{subfigure}[t]{0.16\textwidth}
\includegraphics[width=1\textwidth]{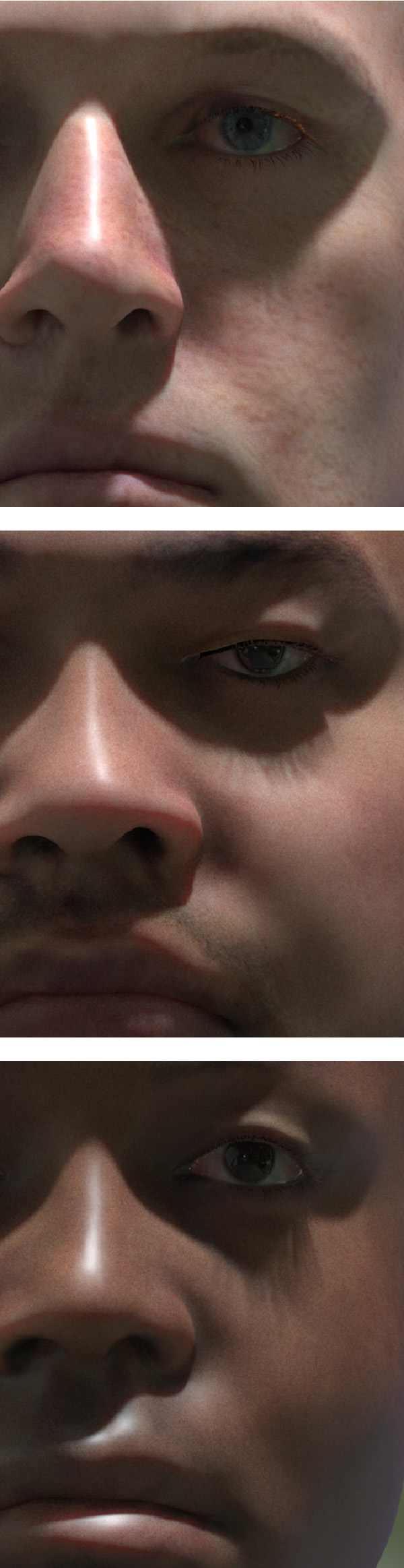}
\caption{}
\end{subfigure}
\begin{subfigure}[t]{0.16\textwidth}
\includegraphics[width=1\textwidth]{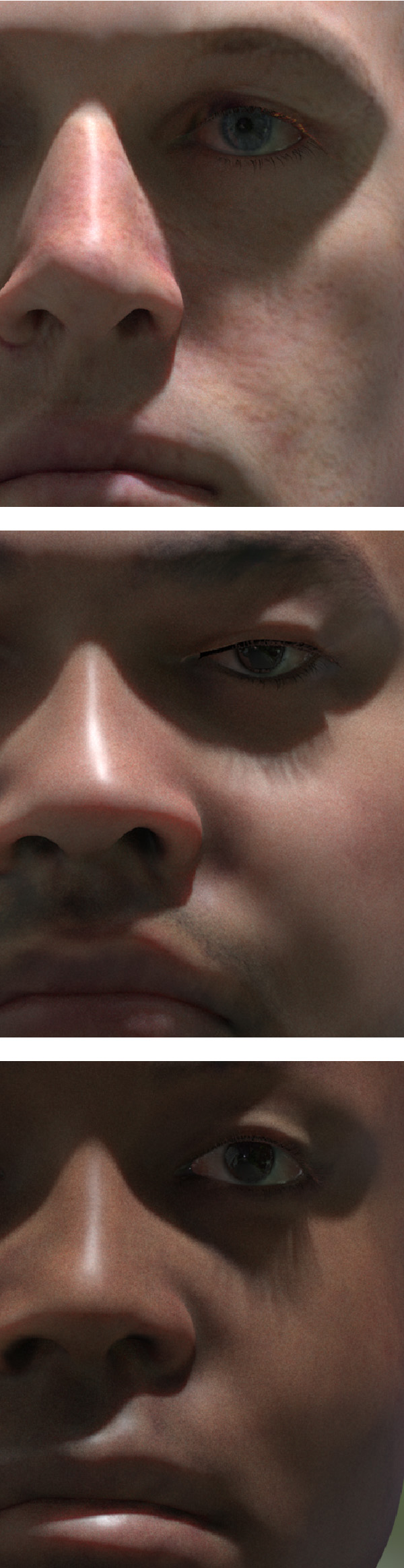}
\caption{}
\end{subfigure}
\begin{subfigure}[t]{0.16\textwidth}
\includegraphics[width=1\textwidth]{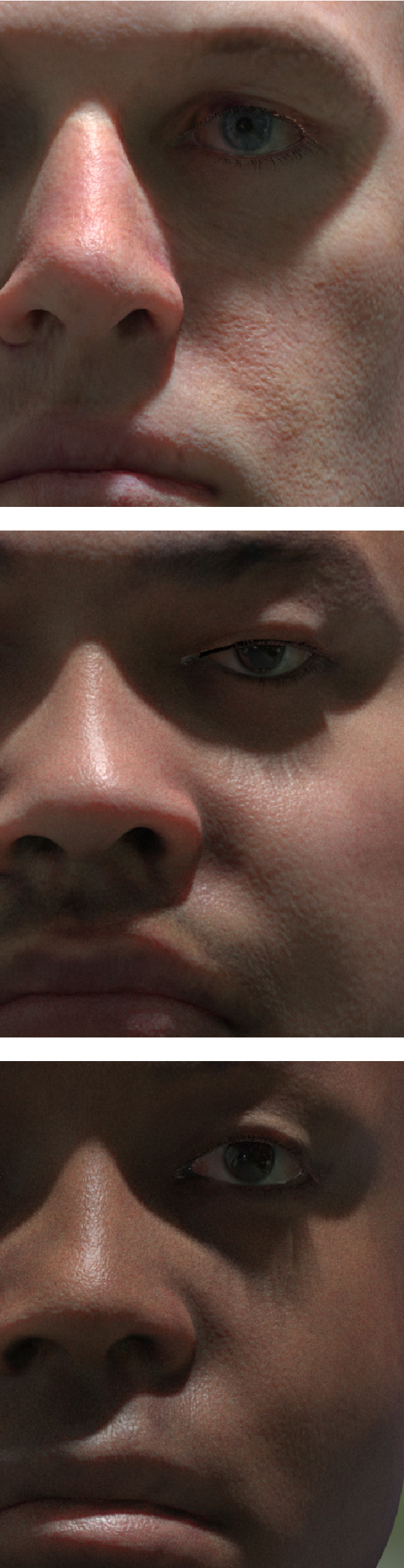}
\caption{}
\end{subfigure}
\end{center}
\caption{Rendered images of generated random samples. Column (a), (b), and (c) show images rendered under novel image-based HDRI lighting ~\cite{HDRIHaven}. Column (c), (d), and (e), show geometry with albedo, specular intensity, and displacement added one at the time.}
\label{fig:quality}
\end{figure*}

\begin{figure}[t]
\begin{center}
\includegraphics[width=0.47\textwidth]{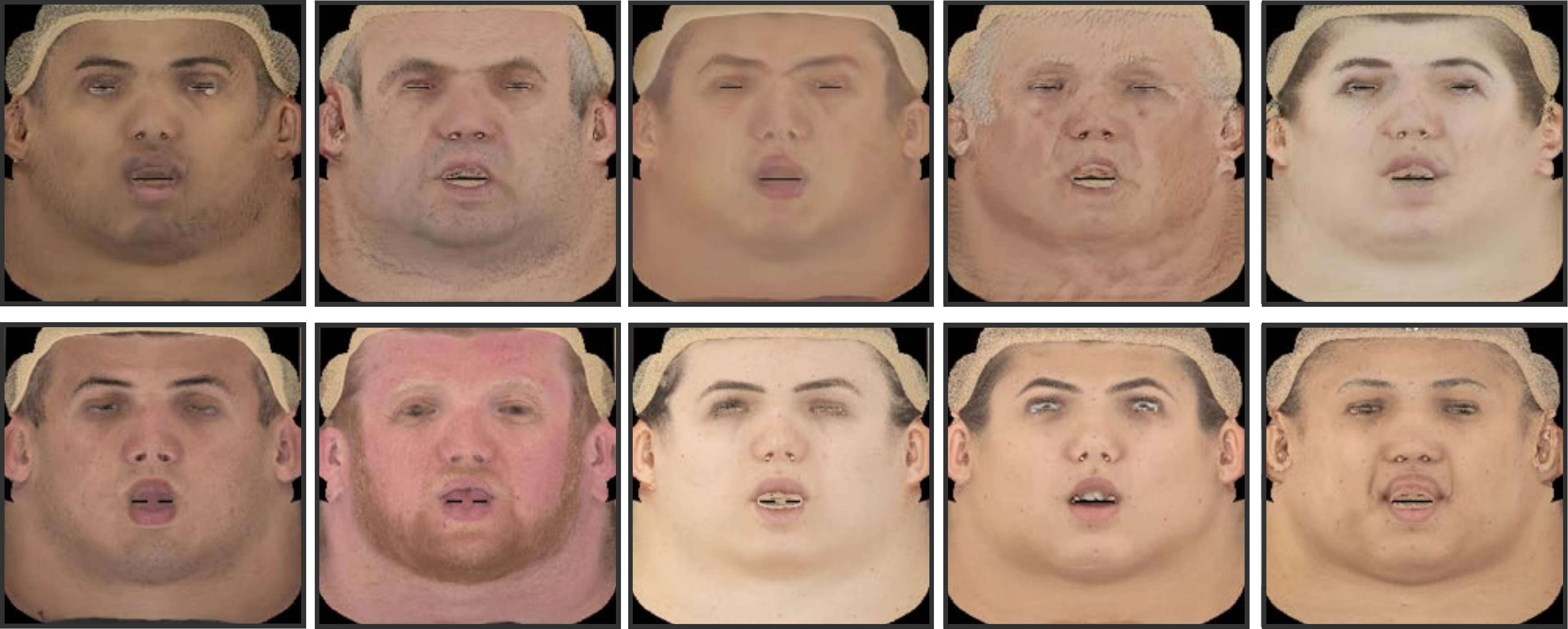}
\vspace{5pt}
\end{center}
   \caption{Nearest neighbors for generated models in training set. Top row: albedo from generated models. Bottom row: albedo of geometrically nearest neighbor in training set.}
\vspace{10pt}
\label{fig:nn_search}
\end{figure}

\begin{table}[t]
\begin{center}
\begin{tabular}{c|c|c}
\toprule[2pt]
Generation Method & IS$\uparrow$ & FID$\downarrow$ \\
\hline\hline
independent & 2.22 & 23.61 \\
joint & \textbf{2.26} & \textbf{21.72} \\
\hline
groud truth & 2.35 & - \\
\bottomrule[2pt]
\end{tabular}
\end{center}
\caption{Evaluation on our \emph{Identity} generation. Both IS and FID are calculated on images rendered with independently/jointly generated albedo and geometry.}
\label{tab:joint}
\end{table}

\subsection{Quantitative Evaluation}

We evaluate the effectiveness of our identity network's \emph{joint} generation in Table \ref{tab:joint} by computing Frechet Inception Distances (FID) and Inception-Scores (IS) on rendered images of three categories: randomly paired albedo and geometry, paired albedo and geometry generated using our model, and ground truth pairs. Based on these results, we conclude that our model generates more plausible faces, similar to those using ground truth data pairs, than random pairing.

We also evaluate our identity network’s generalization to unseen faces by fitting 48 faces from \cite{3DScanstore}. The average Hausdorff distance is $2.8 mm$, which proves that our model's capacity is not limited by the training set.

In addition, to evaluate the non-linearity of our expression network in comparison to the linear expression model of FaceWarehouse \cite{2014FaceWarehouseA3}, we first fit all the Light Stage scans using FaceWarehouse, and get the 25 fitting weights, and expression recoveries, for each scan. We then recover the same expressions by feeding the weights to our expression network. We evaluate the reconstruction loss with mean-square error (MSE) for both FaceWarehouse's and our model's reconstructions. On average, our method's MSE is $1.2 mm$ while FaceWarehouse's is $2.4 mm$. 
This shows that for expression fitting, our non-linear model numerically outperforms a linear model of the same dimensionality.

\begin{figure}[t]
\begin{center}
\begin{subfigure}[b]{0.23\textwidth}
\includegraphics[width=1\textwidth]{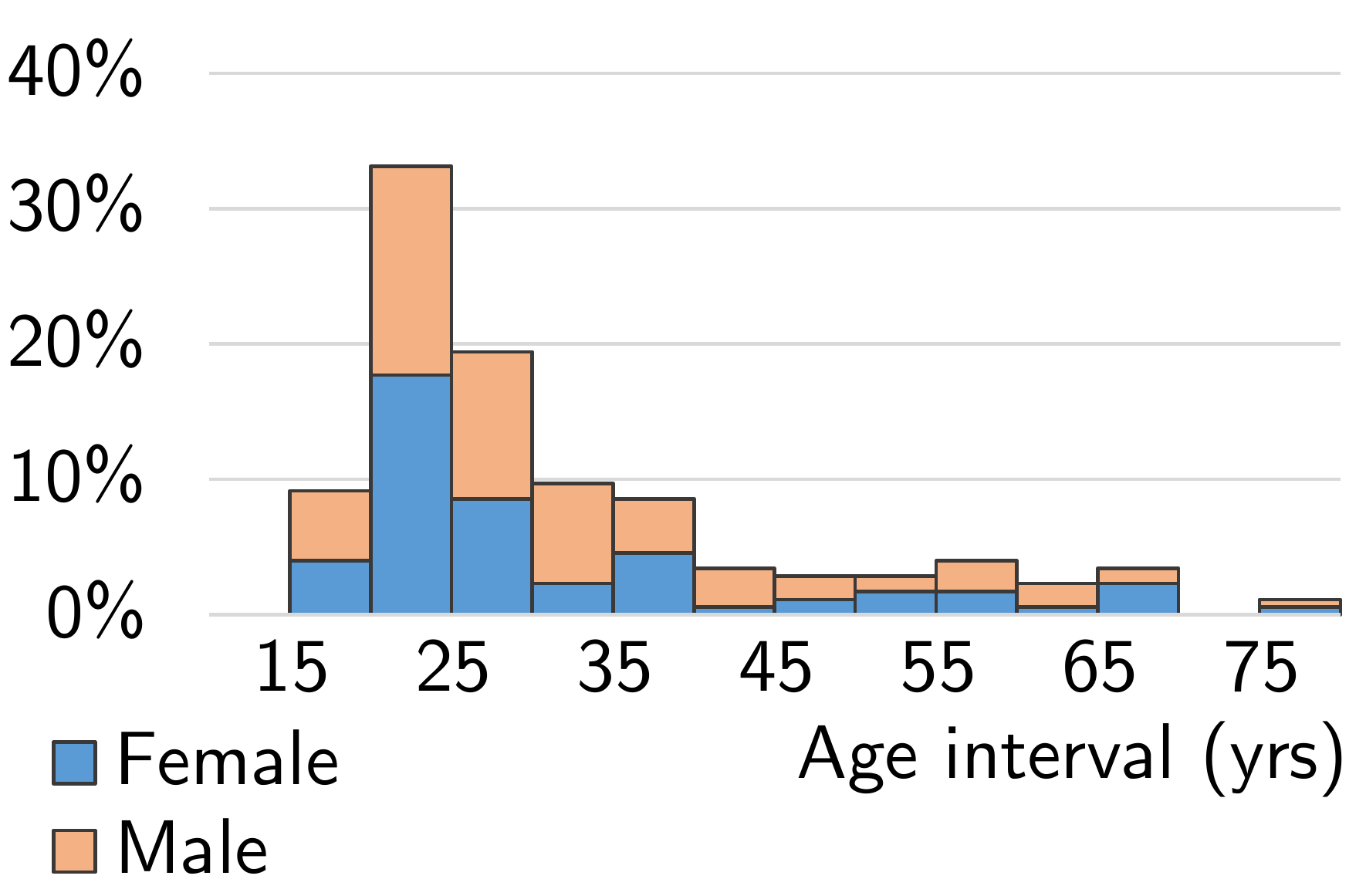}
\caption{Training data}
\end{subfigure}%
\begin{subfigure}[b]{0.23\textwidth}
\includegraphics[width=1\textwidth]{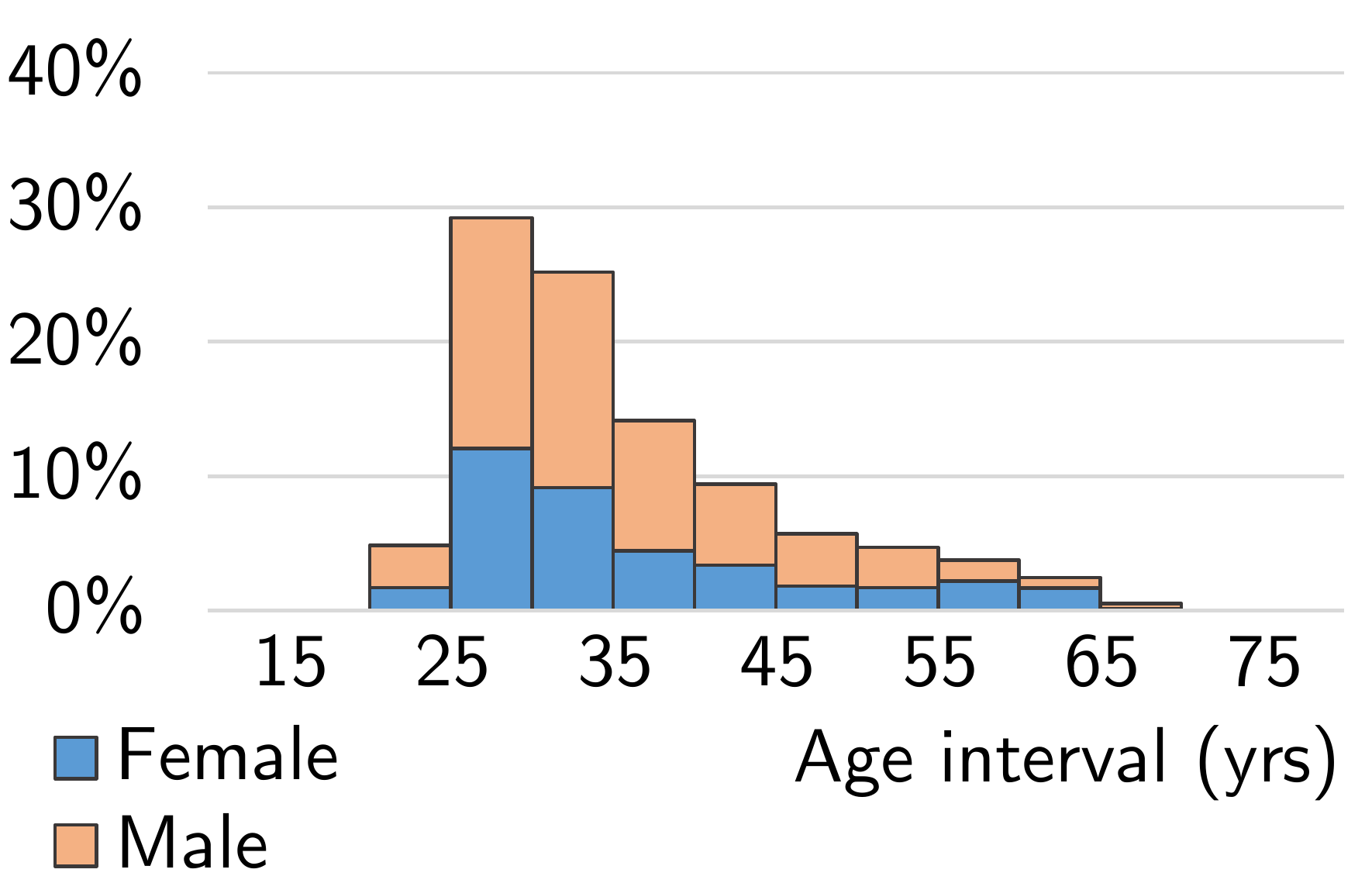}
\caption{Generated data}
\end{subfigure}%
\end{center}
\caption{The age distribution of the training data (a) \emph{VS.} randomly generated samples (b).}
\label{fig:ageDistributionSample}
\end{figure}

To demonstrate our generative identity model's coverage of the training data, we show the gender, and age distributions of the original training data and 5000 randomly generated samples in Fig.\ref{fig:ageDistributionSample}. The generated distributions are well aligned with the source. 

\subsection{Applications}
\begin{figure}[t]
\begin{center}%
\includegraphics[width=0.47\textwidth]{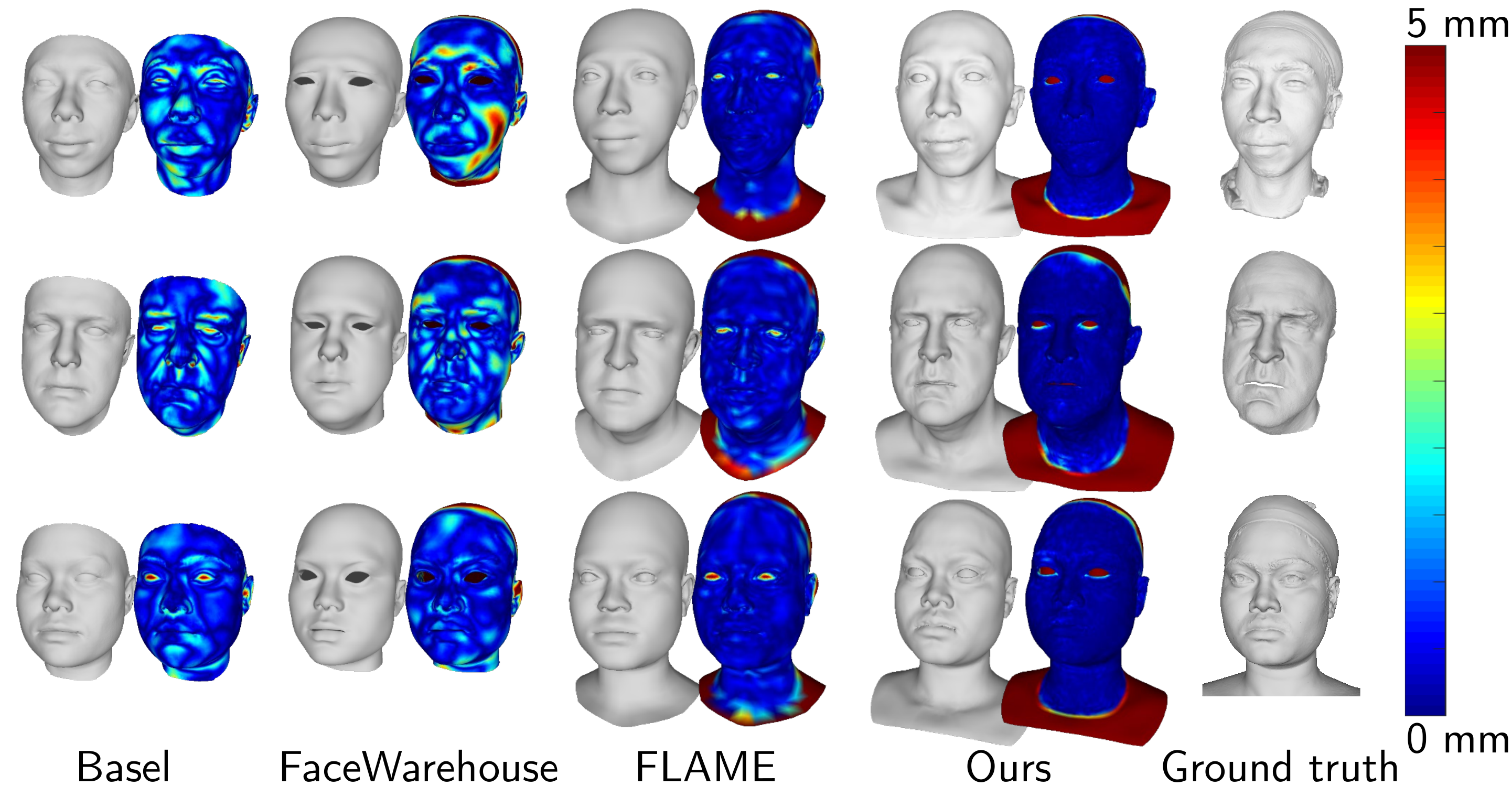}%
\end{center}%
   \caption{Comparison of 3D scan fitting with Basel~\cite{blanz1999morphable}, FacewareHouse~\cite{2014FaceWarehouseA3}, and FLAME~\cite{li2017learning}. Error maps are computed using Hausdorff distance between each fitted model and ground truth scans.}
\label{fig:modelFitting}
\end{figure}

To test the extent of our identity model's parameter space, we apply it to scanned mesh registration by reversing the GAN to fit the latent code of a target image~\cite{lipton2017precise}. As our model requires a 2D parameterized geometry input, we first use our linear model to align the scans using landmarks, and then parameterize it to UV space after Laplacian morphing of the surface. We compare our fitting results with widely used (linear) morphable face models in Fig.\ref{fig:modelFitting}.
This evaluation does not prove the ability to register unconstrained data but shows that our model is able to reconstruct novel faces by the virtue of it's non-linearity, to a degree unobtainable by linear models.

\begin{figure}[t]
\begin{center}
\includegraphics[width=0.47\textwidth]{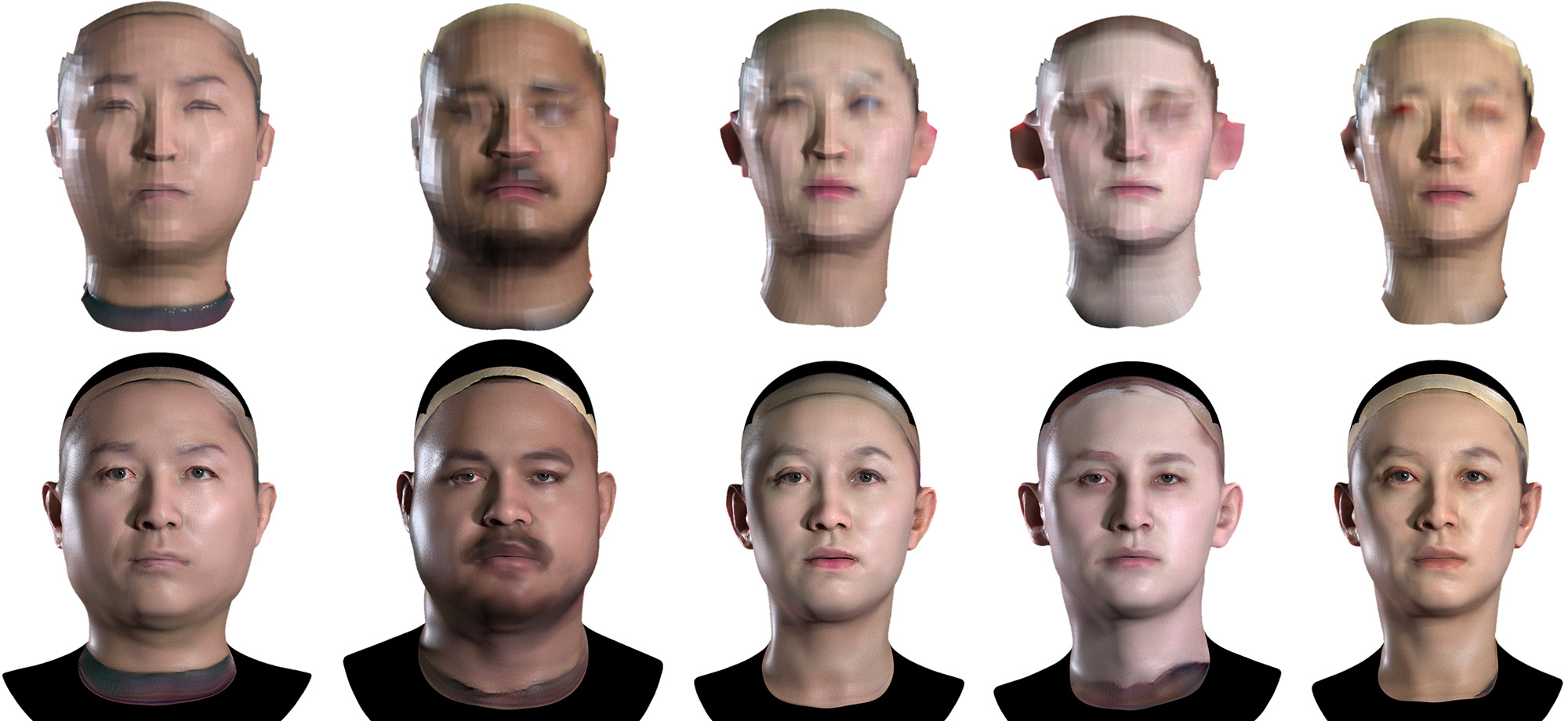}
\end{center}
   \caption{Low-quality data domain transfer. Top row: Models with low resolution geometry and albedo. Bottom row: Enhancement result using our model.}
\label{fig:domainTransfer}
\end{figure}

Another application of our model is transferring low-quality scans into the domain of our model by fitting using both MSE loss and discriminator loss. In Fig.\ref{fig:domainTransfer}, we show examples of data enhancement of low resolution scans.

\section{Conclusion and Limitations}

\paragraph{Conclusion}
We have introduced the first published use of a high-fidelity face database, with physically-based marerial attributes, in generative face modeling. Our model can generate novel subjects and expressions in a controllable manner. We have shown that our generative model performs well on applications such as mesh registration and low resolution data enhancement. We hope that this work will benefit many analysis-by-synthesis research efforts through the provision of higher quality in face image rendering.

\paragraph{Limitations and Future work}
In our model, expression and identity are modeled separately without considering their correlation. Thus the reconstructed expression offset will not include middle-frequency geometry of an individual's expression, as different subjects will have unique representations of the same action unit. Our future work will include modeling of this correlation. Since our expression generation model requires neural network inference and re-sampling of 3D geometry it is not currently as user friendly as blendshape modeling. Its ability to re-target prerecorded animation sequences will have to be tested further to be conclusive.
One issue of our identity model arises in applications that require fitting to 2D imagery, which necessitates an additional differentiable rendering component. A potential problem is fitting lighting in conjunction with shape as complex material models make the problem less tractable. A possible solution could be an image-based relighting method~\cite{sun2019single,Meka:2019} applying a neural network to convert the rendering process to an image manipulation problem.
The model will be continuously updated with new features such as variable eye textures and hair as well as more anatomically relevant components such as skull, jaw, and neck joints by combining data sources through collaborative efforts. To encourage democratization and wide use cases we will explore encryption techniques such as federated learning, homomorphic encryption, and zero knowledge proofs which have the effect of increasing subjects' anonymity.

\section{Acknowledgement}
Hao Li is affiliated with the University of Southern California, the USC Institute for Creative Technologies, and Pinscreen. This research was conducted at USC and was funded by the U.S. Army Research Laboratory (ARL) under contract number W911NF-14-D-0005. This project was not funded by Pinscreen, nor has it been conducted at Pinscreen. The content of the information does not necessarily reflect the position or the policy of the Government, and no official endorsement should be inferred.

{\small
\bibliographystyle{ieee_fullname}
\bibliography{egbib}
}

\newpage

\section*{Appendix}

\section*{Gender Control}
\label{sec:gender}

\paragraph{Step1. Pre-computing mean gender latent code.} First, we propose a classifier $\psi$, trained with ground truth data to classify our input pair (\textit{albedo} and \textit{geometry} maps) into two categories (\textit{male} and \textit{female}). Then we randomly sample $Z_{id} \sim N(\mu_{id}, \sigma_{id})$ to generate $10 k$ sample pairs $G_{id}(Z_{id})$ using our \emph{identity network}.
The classifier separates all the samples into two groups. Finally, we extract the mean vector of each category as $Z_{male}$ and $Z_{female}$ using equation \ref{equ:calMean}.

\begin{equation}
Z_{mean} = \frac{1}{\sum_{i=1}^{10 k} \Omega({Z_{id}^{(i)}})  }\sum_{i=1}^{10 k} Z_{id}^{(i)} \cdot \Omega({Z_{id}^{(i)}})   \\
\label{equ:calMean}
\end{equation}

Where $\Omega({Z_{id}})$ is the gender activation function which converts the outputs of gender classifier $\psi$ into binary values defined as follows:
\begin{align}
\label{equ:genderClassifier}
  \Omega({Z_{id}})=\begin{cases}
               1,   \psi (G_{id}(Z_{id})) <= 0.5\\
               0,   \psi (G_{id}(Z_{id})) > 0.5
            \end{cases}
\end{align}

Where $\Omega({Z_{id}})=1$ is defined to be female, and $\Omega({Z_{id}})=0$ means male. In equation~\ref{equ:calMean}, the mean vector in each category $Z_{male}$ and $Z_{female}$ is computed by simply averaging the samples where $\Omega(Z_{id}^{(i)})$ equals to $1$ and $0$ separately.

\paragraph{Step2. Conditioned Generation.} Instead of directly using a randomly sampled $Z_{id}\sim N(\mu_{id}, \sigma_{id})$ as input, we combine it with the mean gender latent code $Z_{male}$ and $Z_{female}$:
\begin{equation}
    Z_{id}^{gender} = (1 - \alpha - \beta) \times Z_{id} + \alpha \times Z_{male} + \beta \times Z_{female}
\end{equation}

\noindent We can set $\alpha = 0.5, \beta = 0.0$ to ensure generated results are all male, or $\alpha = 0.0, \beta = 0.5$ to ensure generated results are all female. We can also gradually decrease $\alpha$ and increase $\beta$ at the same time to interpolate a male generation into female. An example of this is shown in Fig.9 of the paper.

\section*{Age Control}
The main idea of age control is similar to the gender control (Sec~\ref{sec:gender}) with two main differences: $(1)$ Instead of a classifer $\psi$ for gender classification, we use a regressor $\phi$ to predict the true age (in years). $(2)$ We compute an average vector for $Z_{old}$ and $Z_{young}$ separately using the method of sampling $Z_{id}$ with $\phi (G_{id}(Z_{id})) > 50$ and $\phi (G_{id}(Z_{id})) < 30$. So the final age latent code is represented as:
\begin{equation}
    Z_{id}^{age} = (1 - \alpha - \beta) \times Z_{id} + \alpha \times Z_{old} + \beta \times Z_{young}
\end{equation}

\noindent Figure 9 in the main paper also shows a example of aging interpolation by gradually increasing $\alpha$ from 0.0 to 0.7, and decreasing $\beta$ from 0.7 to 0.0.

\section*{3D Model Fitting}
\label{sec:modelfitting}
Given a face scan, or face model, we firstly convert it into our albedo and geometry map format by fitting a linear face model followed by Laplacian warping and attribute transfer. The ground truth latent code of the input is denoted $Z_{id}$. Our goal of fitting is to find the latent code $Z_{id}^{'}$ that best approximates $Z_{id}$ while retaining the embodyment of our model. To achieve this, one can find $Z_{id}^{'}$ that minimizes $MSE(G_{id}(Z_{id}^{'}), G_{id}(Z_{id}))$ through gradient descent.

In particular, we first use the \textit{Adam} optimizer with a constant $learningrate = 1.0$ to update the input variable $Z_{id}^{'}$, then we update the variables in the \textit{Noise Injection} Layers with $learningrate = 0.01$ to fit those details. Fig.10 in the paper shows the geometry of the fitting results.

\begin{figure}[t!]
\begin{tabular}{cc}
  \includegraphics[width=0.22\textwidth]{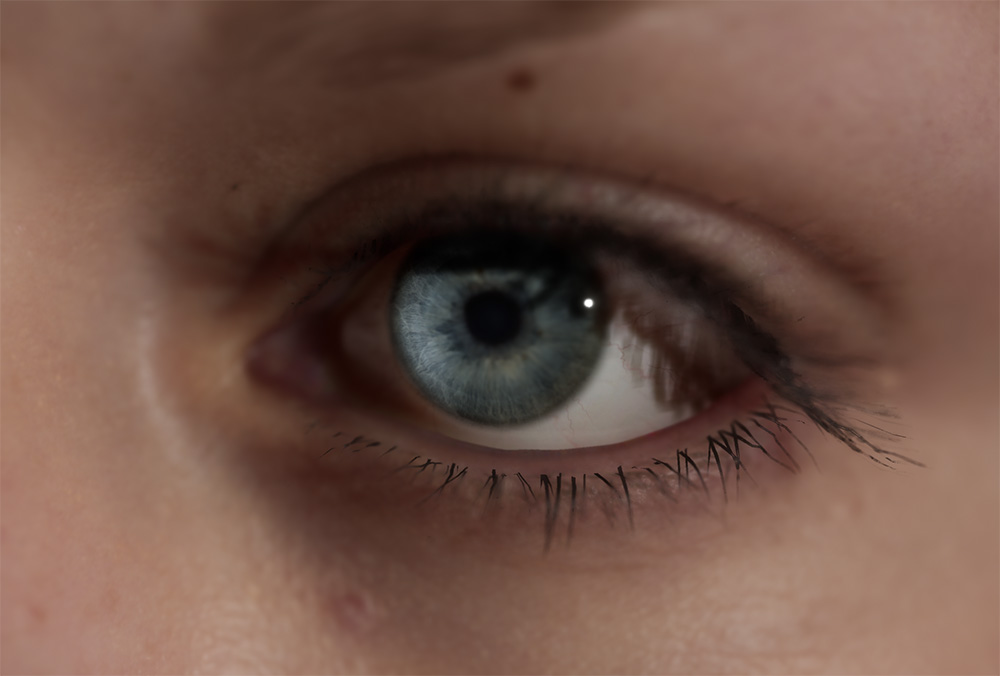} &   \includegraphics[width=0.22\textwidth]{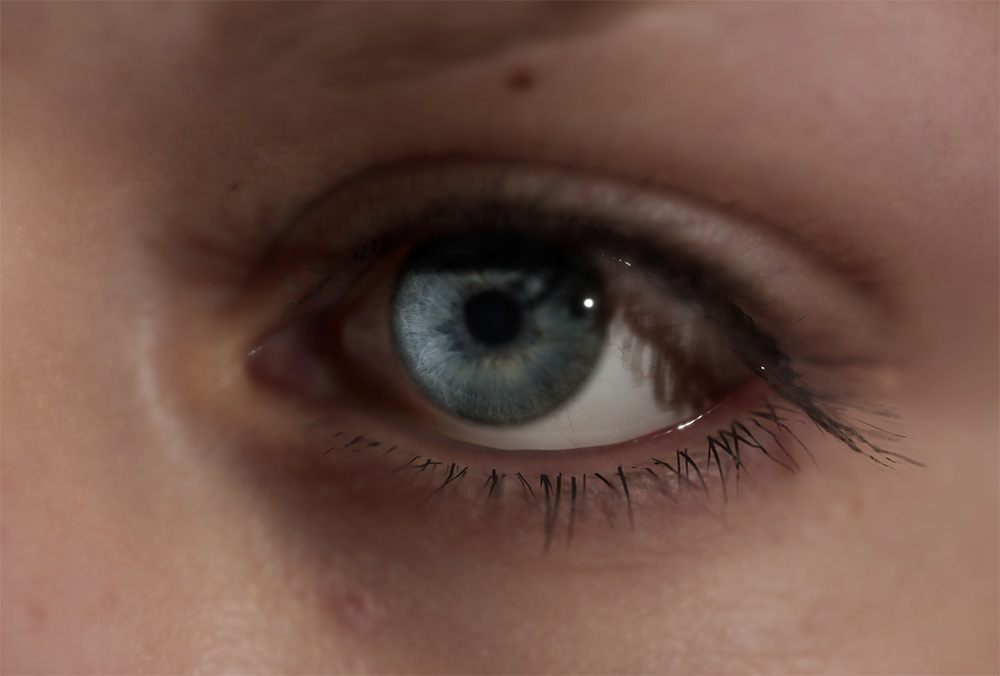} \\
(a) No extra assets & (b) + Lacrimal fluid \\[2pt]
 \includegraphics[width=0.22\textwidth]{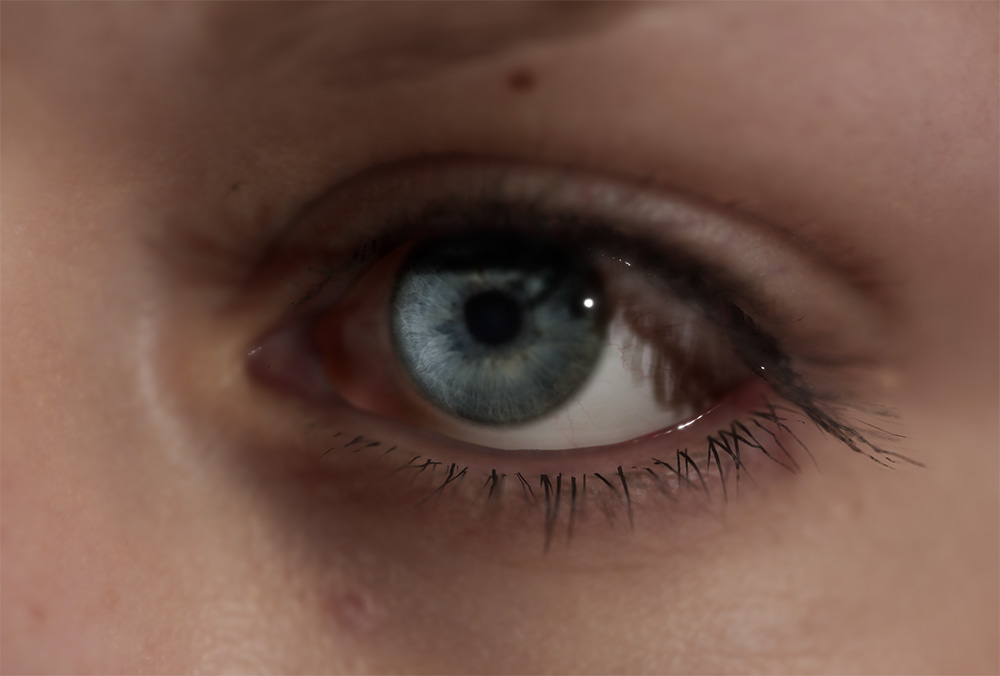} &   \includegraphics[width=0.22\textwidth]{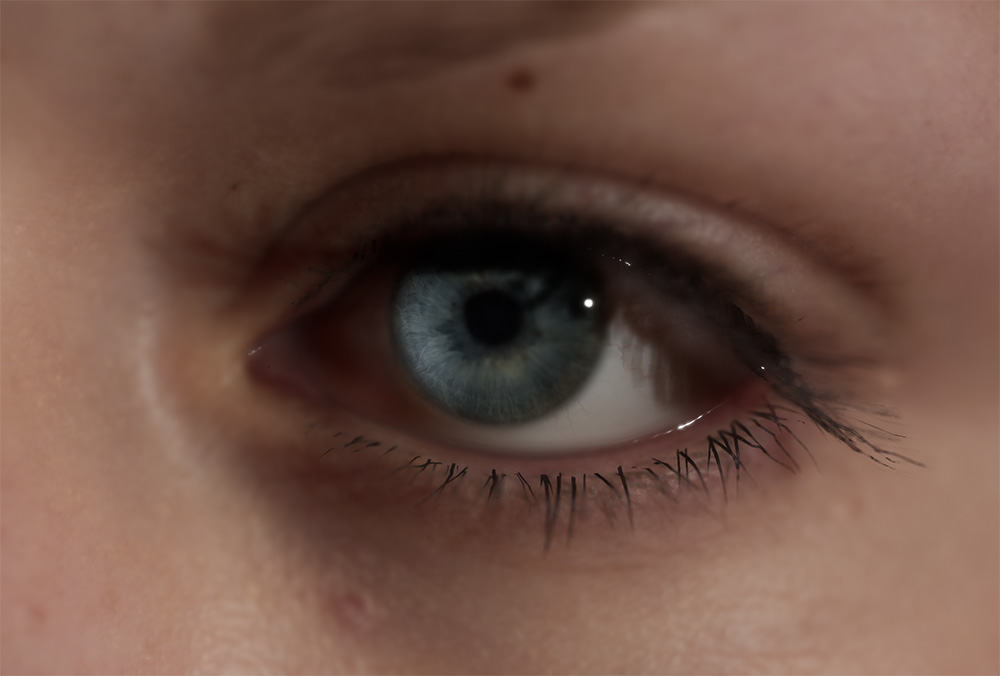} \\
(c) + Blend mesh & (d) + Occlusion mesh \\[2pt]
\end{tabular}
\vspace{10pt}
\caption{Closeup of real time rendered eye with our model's additional eye geometries successively added. The eyeball and eyelashes are considered as default eye geometry and therefore kept in all subfigures.}

\label{fig:eye-assets}
\end{figure}

\section*{Low-quality Data Enhancement.}
In order to enhance the quality of low-resolution data, so that it can be better utilized, the data point needs to be encoded as $Z_{id}$ in our latent space. This is done using our fitting method \ref{sec:modelfitting}. The rest of the high fidelity assets are generated using our generative pipeline. Unlike the fitting procedure, we don't want \textit{true-to-groundtruth} fitting which would result in a recreation of a low resolution model. We instead introduce a discriminator loss to balance the \textit{MSE} loss. This provides an additional constraint on reality and quality during gradient descent. Empirically we give a $0.001$ weight to the discriminator loss to balance the \textit{MSE} loss. We also use the \textit{Adam} optimizer with a constant $learning-rate = 1.0$ for this experiment. The attained variable $Z_{id}^{'}$ is then fed in as the new input, and the process is iteratively repeated until convergence after about 4000 iterations. 

\section*{Real Time Rendering Assets}

To demonstrate the use of additional eye rendering assets (lacrimal fluid, blend mesh, and eye occlusion) available in our model, we show a real time rendering of a close up of an eye and its surrounding skin geometry and material from scan data in Figure \ref{fig:eye-assets}. The rendering is performed using Unreal Engine 4. Materials and shaders are adopted from the \emph{Digital Human} project \cite{UEDigitalHuman}.


\end{document}